\documentclass[runningheads]{llncs}

\usepackage{eccv}

\usepackage{eccvabbrv}

\usepackage{graphicx}
\usepackage{booktabs}
\usepackage{multirow}
\usepackage{colortbl}
\usepackage{capt-of}
\usepackage{xspace}
\usepackage{times}
\usepackage{amsmath}
\usepackage{amssymb}
\usepackage{xpatch}
\usepackage{array}
\usepackage{multirow}
\usepackage{pifont}
\usepackage{enumitem}
\usepackage{bbm}
\usepackage{newtxtext}
\usepackage{algorithm}
\usepackage{algorithmicx}
\usepackage[noend]{algpseudocode}
\usepackage{placeins}
\usepackage{wrapfig}
\usepackage{marvosym}
\usepackage{makecell}

\usepackage{titletoc}

\definecolor{Gray}{gray}{0.9}

\renewcommand{\paragraph}[1]{\vspace{.5em}\noindent\textbf{#1.}}

\usepackage[accsupp]{axessibility}  

\usepackage{hyperref}
\usepackage{bookmark}

\usepackage{orcidlink}

\begin{document}

\title{Optimization-Guided Diffusion for Interactive Scene Generation} 



\author{
Shihao Li\inst{1,2,3}\textsuperscript{\Letter}%
\orcidlink{0000-0002-8094-8701} \and
Naisheng Ye\inst{2}\orcidlink{0009-0008-0574-2763} \and
Tianyu Li\inst{2\dagger}\orcidlink{0009-0008-3838-160X} \and
Kashyap Chitta\inst{4,5}\orcidlink{0000-0002-3891-3230} \and
Tuo An\inst{3}\orcidlink{0009-0005-9730-9761} \and
Peng Su\inst{6} \and
Boyang Wang\inst{1}\orcidlink{0000-0003-3613-8792} \and
Haiou Liu\inst{1}\orcidlink{0009-0001-8341-2566} \and
Chen Lv\inst{3}\orcidlink{0000-0001-6897-4512} \and
Hongyang Li\inst{2}\orcidlink{0000-0001-9110-5534}
}

\authorrunning{S.~Li et al.}


\institute{
Beijing Institute of Technology, Beijing, China \and
OpenDriveLab, The University of Hong Kong, Hong Kong, China \and
Nanyang Technological University, Singapore \and
KE:SAI -- Kyutai ELLIS Scalable Autonomous Intelligence, Tübingen, Germany \and
NVIDIA Research, Germany \and
Yinwang Intelligent Tech. Co. Ltd., Shenzhen, China \\
\email{lishihao.shawn@gmail.com}\\
\url{https://opendrivelab.com/OMEGA}
}

\maketitle

\begin{abstract}
Realistic and diverse multi-agent driving scenes are crucial for evaluating autonomous vehicles, but safety-critical events which are essential for this task are rare and underrepresented in driving datasets. Data-driven scene generation offers a low-cost alternative by synthesizing complex traffic behaviors from existing driving logs. However, existing models often lack controllability or yield samples that violate physical or social constraints, limiting their usability. We present OMEGA, an optimization-guided, training-free framework that enforces structural consistency and interaction awareness during diffusion-based sampling from a scene generation model. OMEGA re-anchors each reverse diffusion step via constrained optimization, steering the generation towards physically plausible and behaviorally coherent trajectories. Building on this framework, we formulate ego–attacker interactions as a game-theoretic optimization in the distribution space, approximating Nash equilibria to generate realistic, safety-critical adversarial scenarios. Experiments on nuPlan and Waymo show that OMEGA improves generation realism, consistency, and controllability, increasing the ratio of physically and behaviorally valid scenes from 32.35\% to 72.27\% for free exploration capabilities, and from 11\% to 80\% for controllability-focused generation. Our approach can also generate $5\times$ more near-collision frames with a time-to-collision under three seconds while maintaining the overall scene realism.
\end{abstract}

\begingroup
\renewcommand\thefootnote{}\footnotetext{
$^\dagger$: Project Lead.\\
}
\endgroup

\begin{figure*}[t]
    \centering
    \includegraphics[width=\linewidth]{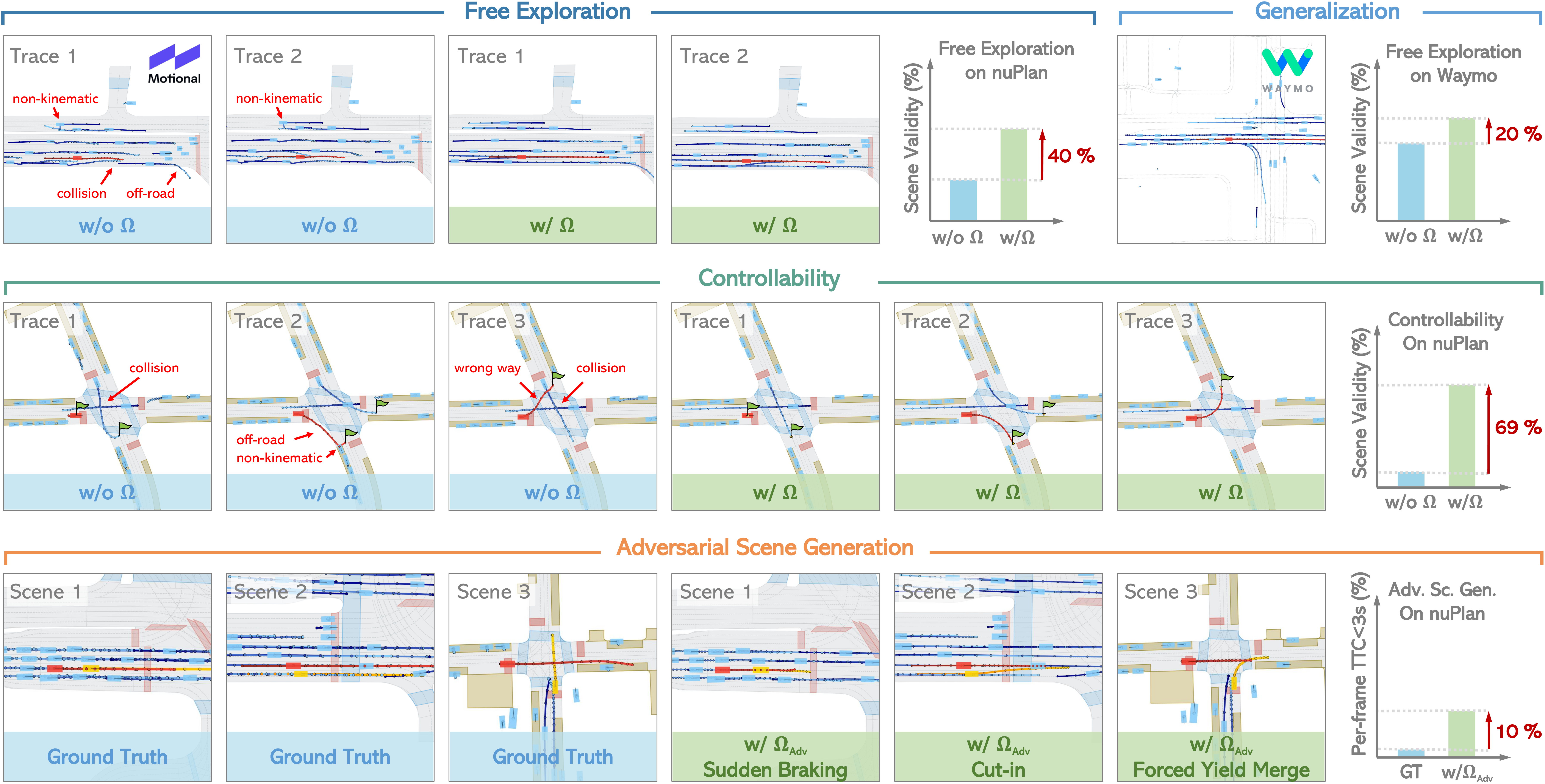}
    \caption{\textbf{OMEGA \((\mathrm{\Omega})\).}
    Our training-free, optimization-guided sampler can be dropped into existing diffusion-based driving scene generators to provide \textit{significantly more realistic, controllable, and interactive} traffic scenarios.
    \textbf{Top.} Our guidance produces diverse and realistic future scenarios from the same historical initialization, increasing the rate of physically and behaviorally valid scenes on both nuPlan (used during the base model's training) and Waymo (zero-shot deployment) compared to unguided baselines.
    \textbf{Middle.} Controllability-conditioned generation synthesizes scenes where vehicles must reach predefined goal points (green flags) for one or multiple target agents, achieving precise behavioral control while maintaining natural interactions.
    \textbf{Bottom.} Finally, we enable adversarial scene generation, adaptively producing diverse attack scenarios without explicitly specifying attacker trajectories.
    Color gradients indicate temporal progression along each trajectory: \textbf{ego} (red gradient), \textbf{attacker} (yellow gradient), and \textbf{other vehicles} (blue gradient).}
    \label{fig:omega_result}
\end{figure*}

\section{Introduction}
\label{sec:intro}

Generating realistic multi-agent driving scenes is essential for the development and evaluation of autonomous vehicles.
However, existing rule-based or physics-based simulators \cite{dosovitskiy2017carla, lopez2018microscopic} fall short: they can reproduce simple traffic patterns but often yield unrealistic or overly scripted interactions. Real-world datasets \cite{caesar2020nuscenes,dauner2024navsim} also offer limited coverage, particularly for safety-critical long-tail events that are crucial for assessing robustness yet occur too infrequently to capture at scale \cite{liu2024curse}. 
These shortcomings highlight the need for a high-fidelity driving scene generator capable of producing realistic, diverse, and controllable multi-agent interactions, especially those involving rare behaviors, which has become an indispensable requirement for efficient and safe evaluation of autonomous driving systems \cite{lin2025model,ma2024unleashing,li2025mtgs,Cao2025CORL,tian2025simscale}.

To meet these requirements, researchers have turned to data-driven scene generation, which enables models to learn complex agent behaviors directly from large-scale driving datasets \cite{hu2024solving, seff2023motionlm, feng2022trafficgen}. Among these, diffusion models have shown remarkable potential in generating realistic multi-agent trajectories and coherent scene layouts, owing to their strong capacity to model high-dimensional, multi-modal distributions and produce samples consistent with empirical data \cite{jiang2024scenediffuser, huang2024versatile}. Despite this success, existing diffusion-based generators face two key limitations.

First, while the learned denoising network implicitly encodes physical and social constraints from data during training, such structure often degrades during inference. As noise is progressively removed, small inconsistencies can accumulate across steps, leading to trajectories that violate kinematic feasibility, social compliance, or interaction logic. These physically or behaviorally inconsistent samples reduce the credibility of generated scenes and limit their utility for downstream safe evaluation.

Second, long-tail and adversarial interactions lie in low-density regions of naturalistic datasets and are therefore underrepresented during training. Consequently, learned models gravitate toward dominant motion patterns and struggle to directly sample rare but safety-critical interactions. Prior diffusion methods introduce guidance via reinforcement-learned classifiers \cite{xie2024advdiffuser}, manually designed behavior-shaping objectives \cite{xu2025diffscene, zhong2022guided}, or pre-specified goal conditions \cite{zhou2025decoupled, jiang2023motiondiffuser}. These strategies either require retraining auxiliary networks or rely on case-by-case guidance functions that demand domain expertise, which limits scalability and generalization across scenarios. 

To address these limitations, we introduce OMEGA, a training-free framework that incorporates structural constraints and strategic intent into diffusion-based driving scene generation.
OMEGA formulates each reverse diffusion step as a conditional sampling process anchored on the estimated clean sample and performs numerical optimization within a KL-bounded trust region to re-anchor the mean of the reverse transition distribution. This optimization-guided formulation progressively steers the reverse Markov chain toward physically consistent and behaviorally coherent trajectories, improving structural fidelity without retraining or altering the backbone diffusion model.

To improve interaction realism, we introduce a phase-aligned guidance schedule that progressively transitions from feasible motion formation (per agent) to reactive coordination (among agents) across two denoising phases. Full-horizon coarse denoising under agent-wise structural priors first establishes a globally plausible motion layout, while fine-grained, time-indexed denoising resolves inter-agent interactions to enhance local reactivity while preserving physical feasibility.

Building on this foundation, we further introduce OMEGA$_\mathrm{Adv}$, a sensitivity-enhanced  adversarial generator
inspired by \cite{fiacco1983introduction,spica2020real}.
It models attacker--ego interactions as a distributional optimization game. The ego agent seeks to maintain natural, feasible behavior, while the attacker, using a sensitivity term that predicts how the ego will react to impending collision-avoidance constraints, seeks to maximally perturb the ego within the learned data distribution.
This formulation enables the adaptive generation of realistic yet challenging driving scenarios without retraining any components or manually prescribing attacker trajectories or target points. Our contributions include:

\begin{itemize}
    \item \textbf{Optimization-guided diffusion sampling.} A training-free method that re-anchors each reverse step via constrained optimization, enforcing structural consistency and stable diffusion guidance.
    \item \textbf{Phase-aligned guidance scheduling.} A stage-wise denoising and guidance scheme that shifts from macro-level plausibility to fine-grained reactivity, improving interaction fidelity and coordinated multi-agent scene evolution.
    \item \textbf{Sensitivity-enhanced adversarial generation.} A game-thoretic formulation for ego--attacker interactions yielding realistic yet challenging scenarios.
    \item  \textbf{Comprehensive empirical results.} OMEGA reduces collisions and off-road rates while improving kinematic feasibility over baselines. It raises scene validity on nuPlan by about 40 points and on Waymo (zero-shot) by about 20 points, and delivers a further 69-point gain in controllability under user-specified intents, mitigating prior validity limitations and bringing diffusion-based scene generation closer to practical integration. It is also capable of generating $5\times$ more near-collision frames under the adversarial scene generation setting.
\end{itemize}

\section{Related Work}
\label{sec:related}

\paragraph{Diffusion models for conditioned generation}
A spectrum of generative architectures has been explored to address driving scenario generation, encompassing autoregressive approaches~\cite{seff2023motionlm,hu2024solving} and diffusion-based frameworks~\cite{jiang2024scenediffuser,tan2025scenediffuser++,zhou2025decoupled,chitta2024sledge}. The latter offers greater flexibility for joint distribution modeling and supports conditional synthesis through guidance mechanisms.
Several guidance strategies have been proposed for trajectory- and scene-level diffusion models.
Some approaches, akin to diffusion forcing~\cite{chen2024diffusion}, interpret Gaussian corruption as continuous partial masking, allowing low-noise tokens to guide reconstruction for macro-level behavior control~\cite{zhou2025decoupled,huang2025mdg} or temporally structured rollouts~\cite{jiang2024scenediffuser}.
While effective for coarse behavioral shaping, such noise modulation provides limited control over fine-grained motion regulation and interaction consistency.
Other approaches introduce guidance during denoising through score or state corrections. CTG~\cite{zhong2022guided,zhong2023language} approximates conditional scores using predicted reverse-step means to enable flexible conditional control, while DPS-style methods~\cite{chung2022diffusion,zheng2025diffusion,huang2026versatile,jiang2023motiondiffuser} further align reverse diffusion updates with clean-space objectives by propagating gradients through the denoiser. However, this guidance relies on local linearization of the nonlinear noise–state mapping and may produce inaccurate update directions. Geometry-constrained approaches such as GHC~\cite{jiang2024scenediffuser} instead enforce feasibility through projection or clipping, introducing discontinuities that may compromise the smoothness and distributional consistency of the generated trajectories.
In contrast, OMEGA integrates noise scheduling and reverse-step guidance in a phase-specific framework, decoupling structural layout formation from fine-grained reactive refinement. By re-anchoring each reverse transition distribution via KL-bounded constrained optimization in the clean space, OMEGA provides more reliable structural control and interaction consistency.

\paragraph{Safety-critical scenario generation}
Generating high-fidelity, safety-critical driving scenarios is paramount for the robust training and evaluation of autonomous driving systems~\cite{Hu2023CVPR,Chen2024PAMI,liu2025reinforced}. Certain approaches employ adversarial learning \cite{zhang2023cat,feng2023dense}, which trains surrounding agents to adversarially interact with the ego vehicle. Gradient-based perturbation methods, including KING~\cite{hanselmann2022king} and AdvSim~\cite{wang2021advsim}, modify background trajectories, or re-simulate sensor data, to induce critical states while preserving physical plausibility. AdvDiffuser \cite{xie2024advdiffuser} injects gradients into the denoising process from an auxiliary collision reward model, steering the diffusion updates toward adversarial trajectories. Nexus~\cite{zhou2025decoupled} leverages corner-case finetuning and manually specified goal-conditioned inpainting to steer selected agents toward predefined outcomes. In contrast to these approaches, our method, OMEGA$_\mathrm{Adv}$, generates scenarios by formulating a distributional game in the optimization guided denoising process, enabling realistic yet safety-critic scenario generation.

\section{Method}
\label{sec:method}

\subsection{Preliminaries}

\paragraph{Diffusion models}
Diffusion models~\cite{sohl2015deep, ho2020denoising, dhariwal2021diffusion, song2021denoising} are likelihood-based generative models that map a simple Gaussian prior to complex data distributions through sequential denoising. 
Given $x_0 \!\sim\! q(x_0)$, the \textit{forward diffusion} adds Gaussian noise with variance $\beta_t \in (0, 1)$ at each step to obtain a sequence of intermediate states $x_1, \ldots, x_T$.
The analytic marginal distribution of conditioned on $x_0$ is:
\begin{equation}
\begin{aligned}
q(x_t \mid x_0)
&= \mathcal{N}\!\big(\sqrt{\bar{\alpha}_t}\,x_0,\; (1-\bar{\alpha}_t)I\big), \\
x_t
&= \sqrt{\bar{\alpha}_t}\,x_0 + \sqrt{1-\bar{\alpha}_t}\,\epsilon.
\end{aligned}
\label{eq:xt_x0}
\end{equation}
where \(\alpha_t = 1 - \beta_t\), \(\bar{\alpha}_t = \prod_{s=1}^t \alpha_s\), and \(\epsilon \sim \mathcal{N}(0, I)\).

Starting from pure noise $x_T \!\sim\! \mathcal{N}(0,I)$, the \textit{reverse process} reconstructs data from noise and is modeled as a Gaussian transition:
\begin{equation}
p_\theta(x_{t-1}\mid x_t)
= \mathcal{N}\!\big(\mu_\theta(x_t, t),\, \Sigma_\theta(x_t, t)\big),
\end{equation}
where in DDPMs~\cite{ho2020denoising} the variance $\Sigma_\theta$ is fixed as $\sigma_t^2 I$, 
and the network learns to predict the noise term $\epsilon_\theta(x_t, t)$ that perturbed $x_0$ to obtain $x_t$.
After training, samples are generated by iteratively applying this reverse transition:
\begin{equation}
x_{t-1}
= \frac{1}{\sqrt{\alpha_t}}
\!\left(x_t - \frac{1 - \alpha_t}{\sqrt{1 - \bar{\alpha}_t}}\,
\epsilon_\theta(x_t, t)\!\right)
+ \sigma_t z,
\label{eq:ddpm_sample}
\end{equation}
where $z \sim \mathcal{N}(0, I)$ is standard Gaussian noise and $t = T, \ldots, 1$. Here, $\beta_t$ denotes the noise schedule, $\alpha_t = 1 - \beta_t$ is the retained signal coefficient, and $\bar{\alpha}_t$ is its cumulative product controlling the noise level.

\paragraph{Problem formulation}
We represent a traffic scene as a spatiotemporal tensor \( x \in \mathbb{R}^{A \times \mathcal{T} \times D} \)
where \( A \) denotes the number of interacting agents, \( \mathcal{T} \) the total number of physical timesteps, and \( D \) the state dimension including position, heading, velocity, and physical size. 
A binary validity mask \( m \in \mathbb{B}^{A \times \mathcal{T}} \) indicates each agent’s existence over time, accounting for dynamic entries and exits.
Following prior works \cite{jiang2024scenediffuser, zhou2025decoupled}, we formulate driving scene generation as a multi-agent inpainting problem over $x$.
Given an inpainting mask \( \tilde{m} \in \mathbb{B}^{A \times \mathcal{T} \times D} \) and its corresponding context values \( \bar{x} = \tilde{m} \odot x \), the diffusion model learns to reconstruct the unobserved elements of \( x \) conditioned on both local spatiotemporal and global scene information.
The given context may include arbitrary known elements of $x$, such as all agents’ past trajectories or designated future goals for specific agents. 
The effective target region for generation corresponds to valid but unobserved elements \( m \odot (1-\tilde{m}) \), representing future state of existing agents or newly introduced ones. Within this formulation, the diffusion model iteratively refines the scene tensor through denoising, progressively completing the partially observed scene toward a coherent multi-agent configuration.

\subsection{Optimization-Guided Diffusion Sampling}
\label{sec:omega}

\paragraph{Anchored reverse transition}
In the standard diffusion sampling process, the model predicts the injected noise $\epsilon_\theta(x_t, t)$ at each step, from which an approximate clean estimate corresponding to the noisy state \(x_t\) can be obtained via \cref{eq:xt_x0}:
\begin{equation}
\tilde{x}_0 = \frac{x_t - \sqrt{1-\bar{\alpha}_t}\,\epsilon_\theta(x_t, t)}{\sqrt{\bar{\alpha}_t}}.
\label{eq:x0_estimate}
\end{equation}
Substituting $\tilde{x}_0$ into the reverse kernel, the posterior distribution of $x_{t-1}$ given $x_t$ can be expressed as:
\begin{equation}
p_\theta(x_{t-1} \mid x_t)
= \mathcal{N}\!\big(x_{t-1};\, A_t \tilde{x}_0 + C_t x_t,\, \sigma_t^2 I\big),
\label{eq:reverse_kernel}
\end{equation}
where $A_t = \frac{(1 - \alpha_t)\sqrt{\bar{\alpha}_{t-1}}}{1 - \bar{\alpha}_t}$ and 
$C_t = \frac{\sqrt{\alpha_t}(1 - \bar{\alpha}_{t-1})}{1 - \bar{\alpha}_t}$ 
are deterministic coefficients determined by the variance schedule.  
Since $x_t$, $A_t$, and $C_t$ are fixed at each step, the conditional distribution of $x_{t-1}$ is fully characterized by the model-estimated clean sample $\tilde{x}_0$.  
For convenience, we denote this Gaussian transition as \( Q_t(\tilde{x}_0) \),  
indicating that the reverse kernel is anchored on $\tilde{x}_0$ while its variance and linear coefficients are fixed by $(x_t, t)$.

Accordingly, \cref{eq:ddpm_sample} can be rewritten as:
\begin{equation}
x_{t-1}
= \underbrace{A_t \tilde{x}_0}_{\text{regression term}}
+ \underbrace{C_t x_t}_{\text{inertia term}}
+ \underbrace{\sigma_t z}_{\text{noise term}}.
\end{equation}
This update can be decomposed into three interpretable components:
(i) a \textbf{regression term} ($A_t \tilde{x}_0$) that pulls the sample toward the model-predicted clean state,
(ii) an \textbf{inertia term} ($C_t x_t$) that preserves continuity with the current noisy state and stabilizes transitions across steps,
and (iii) a \textbf{noise term} ($\sigma_t z$) that maintains sample diversity through controlled stochasticity.
Together, these components describe a reverse process that iteratively refines \(x_t\) toward the predicted clean state \(\tilde{x}_0\), ultimately yielding a sample consistent with the learned data distribution.
This perspective highlights that each reverse diffusion step constitutes a \textit{conditional sampling process anchored on the predicted clean sample}, with \(\tilde{x}_0\) serving as the denoising target that guides the evolution of the Markov chain.


\begin{figure*}[t]
    \centering
    \includegraphics[width=\linewidth]{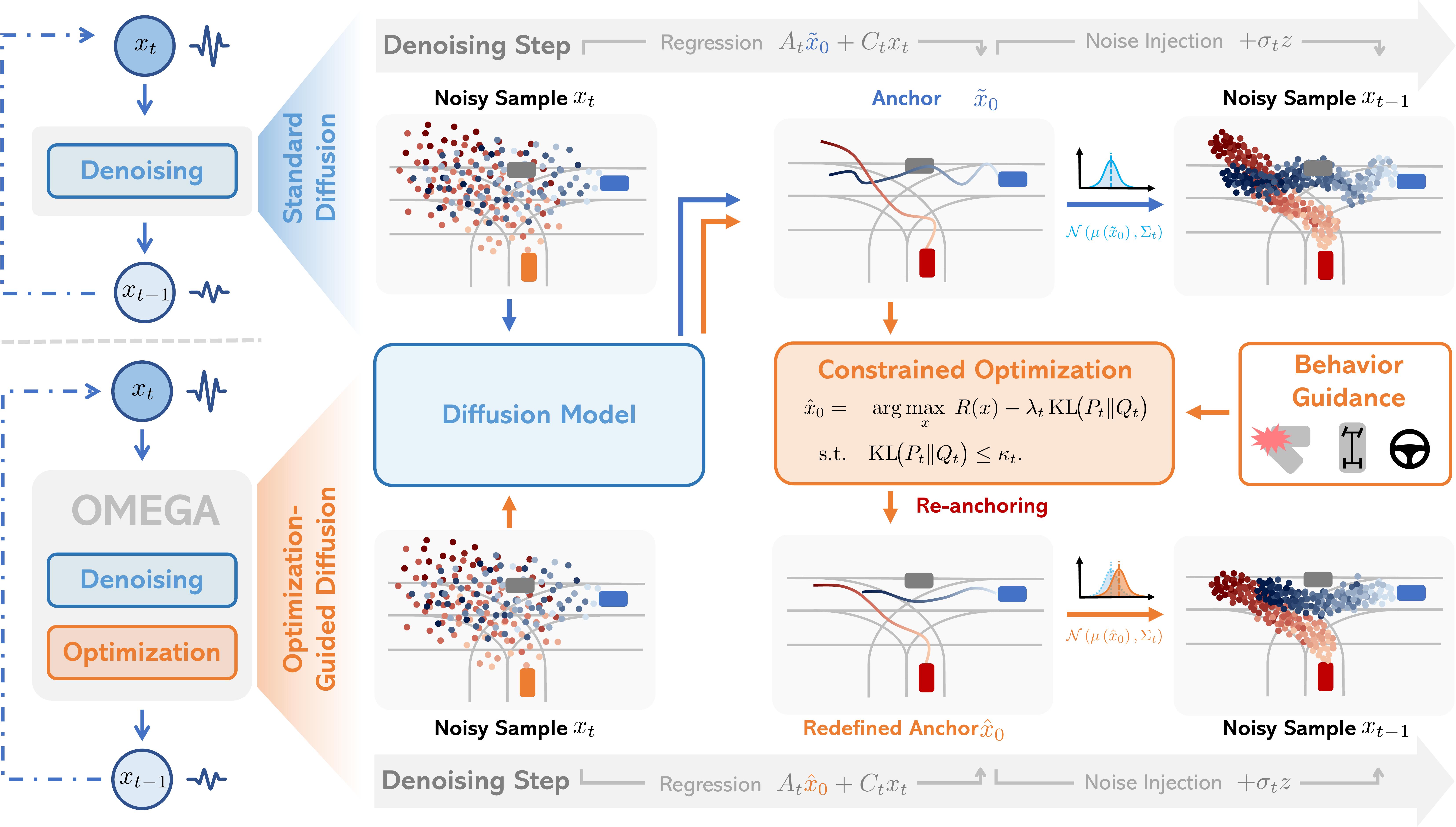}
    \caption{\textbf{Standard diffusion sampling (top) vs. our optimization-guided OMEGA (bottom).}  
    We re-anchor each reverse diffusion step via constrained optimization within a KL-bounded trust region, steering the Markov chain towards behaviorally coherent samples.}
    \label{fig:omega}
\end{figure*}

\paragraph{Optimization-guided re-anchoring}
While the predicted clean sample $\tilde{x}_0$ defines the local denoising direction inferred from the learned dynamics, it is derived purely through data-driven regression and may therefore violate structural or behavioral priors, gradually drifting away from the manifold of physically and behaviorally plausible samples.

\begin{figure*}[!t]
    \centering
    \includegraphics[width=\linewidth]{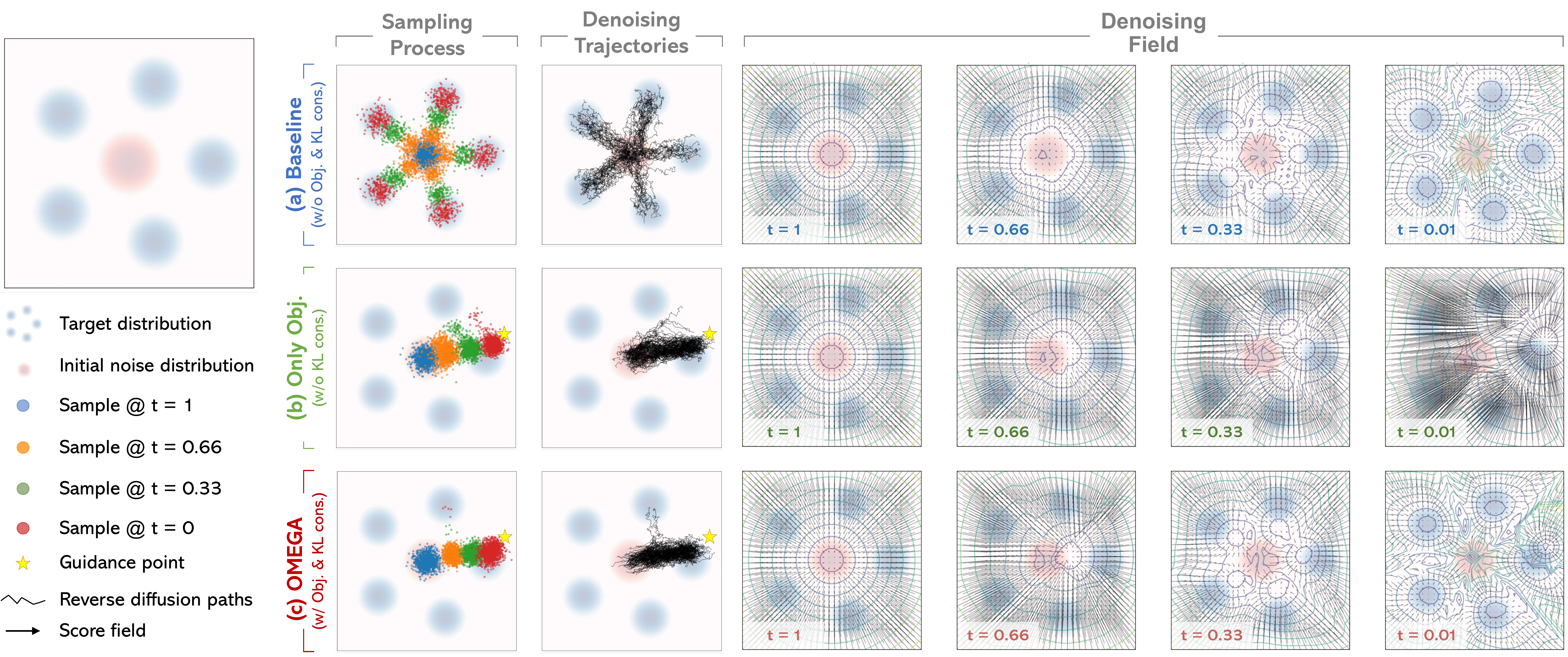}
    \caption{\textbf{Toy example illustrating the effect of OMEGA on generative sampling dynamics.}  
    We adopt a simple two-dimensional generation setup from~\cite{holderrieth2025introduction} to visualize how our method influences the sampling process.
    (a) Without guidance, the diffusion model approximates the target data distribution and recovers its multimodal structure.
    (b) Using only the guidance objective \(R(x)\) pulls samples excessively toward the guidance point, thereby distorting the local sampling geometry.
    (c) Combining the guidance objective with a KL constraint yields controlled adaptation toward the guidance point while suppressing excessive distributional drift.
    Details are provided in the supplementary material.
}
    \label{fig:toy_exm}
\end{figure*}

To explicitly regularize this evolution, OMEGA introduces an optimization-guided refinement of the anchor (Fig.~\ref{fig:omega}), redefining each reverse transition as a new distribution $P_t(\hat{x}_0)$ whose mean is adaptively adjusted within a bounded divergence from the original reverse kernel:
\begin{equation}
\begin{aligned}
P_t(\hat{x}_0) &= \mathcal{N}(A_t \hat{x}_0 + C_t x_t,\, \sigma_t^2 I), \\
\text{s.t.} \quad 
&\mathrm{KL}\!\big(P_t(\hat{x}_0) \,\Vert\, Q_t(\tilde{x}_0)\big) \le \kappa_t.
\end{aligned}
\end{equation}
The optimized anchor $\hat{x}_0$ is obtained by solving a constrained maximization problem:
\begin{equation}
\begin{aligned}
\hat{x}_0
= \arg\max_{x}\;
&\Big[
\lambda_t R(x)
- \, \mathrm{KL}\!\big(P_t(x) \,\Vert\, Q_t(\tilde{x}_0)\big)
\Big], \\
\text{s.t.}\quad
&\mathrm{KL}\!\big(P_t(x) \,\Vert\, Q_t(\tilde{x}_0)\big)
\le \kappa_t.
\end{aligned}
\label{eq:omega_opt}
\end{equation}
Here, \(R(x)\) serves as a differentiable objective that encodes structural and behavioral preferences guiding the generation process, while the KL term defines a trust region that regularizes deviation from the pretrained reverse transition.
As illustrated by the toy example in Fig.~\ref{fig:toy_exm}, this constraint prevents excessive drift or mode collapse, stabilizing the denoising dynamics and preserving consistency with the learned distribution.
Through this balance, the sampler implicitly shifts the mean of the reverse transition toward constraint-consistent regions, adapting its trajectory while maintaining distributional coherence throughout the reverse process.

OMEGA guides the denoising process through distributional re-anchoring, softly steering each reverse step toward constraint-consistent regions while limiting per-step drift and mitigating intermediate-step collapse. By iteratively optimizing the anchor $\hat{x_0}$ and resampling from $P\left(\hat{x}_0\right)$, the process progressively re-aligns the reverse Markov chain with both the model’s learned data distribution and the imposed structural priors, enabling structural yet distributionally coherent generation.

\paragraph{Equivalent euclidean formulation}
Since the anchor $\hat{x}_0$ directly corresponds to the joint future trajectories of all agents in the scene, it admits a physically meaningful interpretation in the optimization space. We represent \(R(x)\) as a structured objective \(r(x)\) equipped with equality and inequality constraints \(h(x)=0\) and \(g(x)\le0\), encoding different levels of structural and behavioral priors. 
Here, \(r(x)\) captures trajectory-level preferences such as smoothness, 
\(h(x)=0\) enforces equality relations such as kinematic feasibility,  
and \(g(x)\le0\) defines inequality conditions such as collision avoidance or boundary adherence.

By substituting the Gaussian forms of \(P_t(x)\) and \(Q_t(\tilde{x}_0)\) into \cref{eq:omega_opt},  
the optimization can be explicitly written in the Euclidean domain as:
\begin{equation}
\begin{aligned}
\hat{x}_0
&=\arg\max_x\;
\lambda_t\, r(x)
-\frac{A_t}{2\sigma_t^2}\,\|x-\tilde{x}_0\|_2^2, \\
&\text{s.t.}\;\;
\|x-\tilde{x}_0\|_2
\le
\sqrt{2\kappa_t}\,\frac{\sigma_t}{|A_t|}, \\
&\qquad h(x)=0,\quad g(x)\le0.
\end{aligned}
\label{eq:euclid-form}
\end{equation}

This equivalent formulation shows that the KL-divergence constraint in \cref{eq:omega_opt} translates into an adaptive Euclidean trust region,  
whose radius scales proportionally with the diffusion variance \(\sigma_t\). The trust region bounds the magnitude of each refinement step, while the guidance signal shapes its direction according to the local scene context. Consequently, early reverse steps with higher stochastic uncertainty allow the optimization to explore a broader set of feasible configurations, whereas subsequent steps at lower variance confine updates near the learned data manifold, thereby promoting stable convergence and preserving consistency with the learned generative dynamics.

\subsection{Phase-Aligned Guidance Schedule}
\label{sec:noise_schedule}
Realistic multi-agent scene generation requires jointly achieving globally plausible motion evolution and fine-grained local reactivity. This entails satisfying heterogeneous objectives for agent-wise motion feasibility and inter-agent coordination. Uniformly enforcing these objectives throughout denoising essentially casts the reverse process as solving a high-dimensional, tightly coupled problem over multiple agents and time steps. More critically, before basic feasible motion structure is established, interaction objectives can be ill-conditioned due to unreliable relative configurations across agents. Enforcing them jointly with objectives for motion feasibility in this regime can therefore yield conflicting guidance signals and unnecessary optimization overhead.

\begin{figure}[t]
    \centering
    \includegraphics[width=1.0\linewidth]{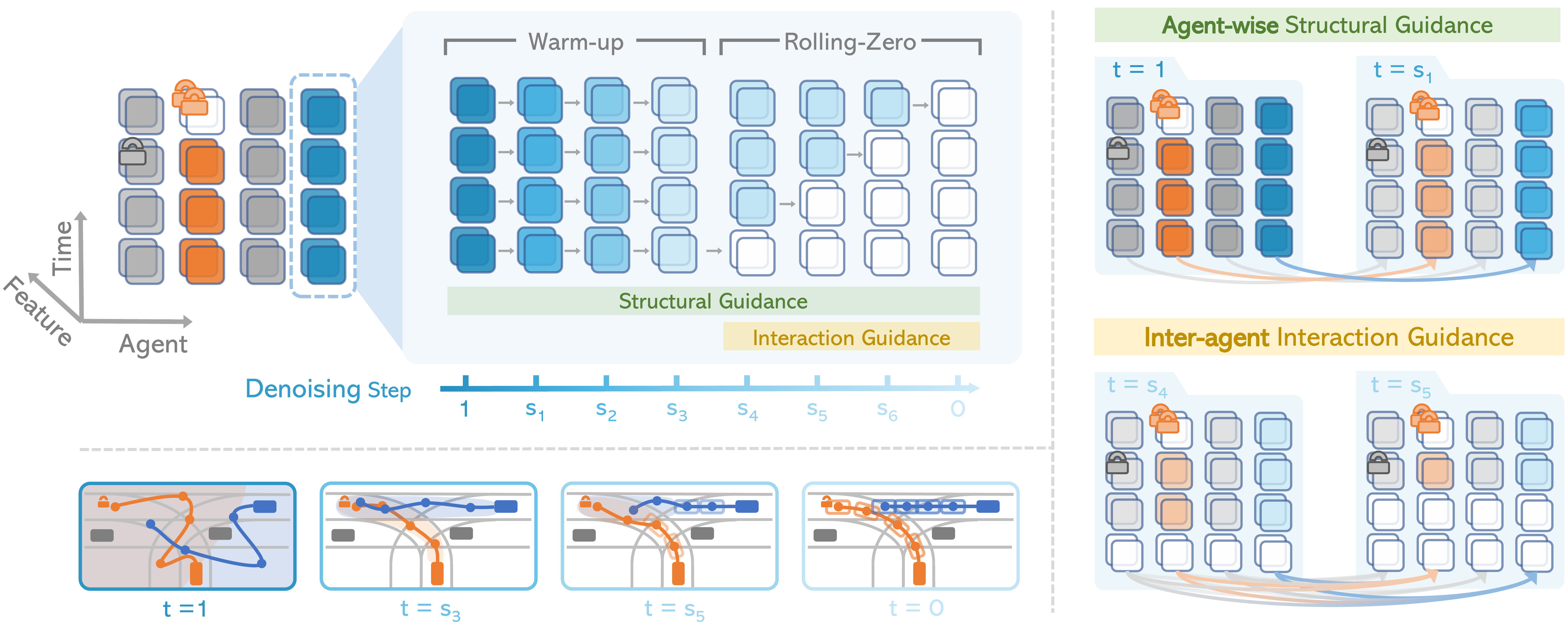}
    \caption{\textbf{Illustration of phase-aligned guidance schedule.} Darker blue means higher noise, lighter blue means lower noise. The green and yellow bars denote phase-specific guidance activation: agent-wise structural guidance is applied throughout denoising to refine each agent under its own motion and scene constraints, while inter-agent interaction guidance is additionally activated during Rolling-Zero to condition updates on the latest denoised states of other agents.}
    \label{fig:noise_schedule}
\end{figure}

To address this issue, we propose a phase-aligned guidance schedule, which integrates a two-phase denoising schedule with phase-specific guidance, as shown in \cref{fig:noise_schedule}. The denoising schedule, consisting of \textit{Warm-up} and \textit{Rolling-Zero}, decomposes scene generation into macro-level layout formation and fine-grained, time-indexed interaction refinement. The guidance objectives are then activated in a phase-aligned manner to reinforce their distinct roles in feasible motion formation and reactive coordination.

\paragraph{Warm-up phase}
The Warm-up phase starts from the fully corrupted scene state and performs full-sequence denoising over the entire future horizon under a gradually decreasing noise schedule. This global optimization establishes macro-level spatial organization and dynamic plausibility, ensuring smooth and kinematically coherent motion throughout the horizon. By treating the temporal sequence as a unified denoising target, Warm-up suppresses autoregressive drift and stabilizes long-horizon trends. It then halts in a low-noise regime, retaining controlled residual uncertainty and providing a well-posed initialization for subsequent refinement.

\paragraph{Rolling-Zero phase}
Operating in the low-noise regime produced by Warm-up, the Rolling-Zero phase progressively removes the residual noise through a time-indexed denoising schedule. At each step, one frame is refined conditioned on the progressively current denoised history and partially masked future states. This temporally localized process enables fine-grained adaptation to evolving inter-agent interactions while preserving the globally coherent structure established in Warmup, thereby gradually restoring frame-level responsiveness.

\paragraph{Phase-specific guidance}
Guidance is scheduled across the two phases. During Warm-up, it is restricted to agent-wise structural objectives, including motion feasibility, road compliance, state consistency, and smoothness, to form a globally plausible motion layout and coarse behavior intent. During Rolling-Zero, inter-agent interaction guidance is activated in a time-indexed manner and evaluated on the latest denoised scene state, enabling timely responses to evolving multi-agent motions. This phase-aligned design reduces joint optimization burden and naturally supports parallel computation.

\subsection{Sensitivity-Enhanced Adversarial Generation}
\label{sec:omega_se}

Building upon the OMEGA formulation, OMEGA$_\mathrm{Adv}$ extends it to adversarial scene generation by formulating a sensitivity-enhanced distributional game between the ego vehicle \(e\) and an attacker \(a\). While the ego aims to preserve feasibility and realism within the learned generative manifold, the attacker seeks to induce maximal perturbations that remain distributionally consistent, thereby exposing safety-critical yet physically plausible interactions. This formulation enables adversarial scenario synthesis directly within the learned generative space, without retraining or external trajectory scripting.

Following this distributional game setup, both participants perform constrained optimization: the ego strives to maintain realistic feasibility:
\begin{equation}
\begin{aligned}
\hat{x}_0^e = 
&\arg\max_{x^e} \; J^e(x^e) \\
\text{s.t.}\quad
&\|x^e - \tilde{x}_0^e\|_2 \le \rho_t, \;
h(x^e)=0,\; g(x^e)\le0, \\
&\gamma_e(x^e,x^a)\le 0,
\end{aligned}
\label{eq:ego_opt}
\end{equation}
while the attacker perturbs ego behavior by solving:
\begin{equation}
\begin{aligned}
\hat{x}_0^a =
&\arg\max_{x^a} \; J^a(x^a) - \alpha J^e(x^e) \\
\text{s.t.}\quad
&\|x^a - \tilde{x}_0^a\|_2 \le \rho_t, \;
h(x^a)=0,\; g(x^a)\le0, \\
&\gamma_a(x^a,x^e)\le 0.
\end{aligned}
\label{eq:attacker_opt}
\end{equation}
Here, \(J^e\) and \(J^a\) share the same structural form as \cref{eq:euclid-form}, 
promoting consistency with the learned generative manifold, 
while \(\alpha>0\) balances the attacker’s aggressiveness against realism.
The pairwise safety constraint \(\gamma(x^e,x^a)\) couples the two subproblems,
inducing a noncooperative differential game where each player’s feasible set depends on the other’s strategy and is activated only within their respective responsibility regions.

\paragraph{Sensitivity-enhanced iterative best response}
Directly solving the joint optimization in Eqs. (\ref{eq:ego_opt})-(\ref{eq:attacker_opt}), which constitutes a noncooperative differential game, is computationally prohibitive. To approximate its Nash equilibrium efficiently, we adopt a sensitivity-enhanced iterative best response (SE-IBR) scheme \cite{spica2020real}. Starting from the model-predicted anchors \(\tilde{x}_0^e\) and \(\tilde{x}_0^a\) as initial strategies, the ego and attacker alternately update their clean anchors as best responses to each other, iterating until convergence.  

Let \(x^{a(l-1)}\) denote the attacker’s previous anchor and \(x^{e(l)}\) the ego’s current optimized response. 
Since the ego’s optimal value \(J^{e*}(x^a)\) implicitly depends on the attacker’s decision \(x^a\),  SE-IBR characterizes its local sensitivity around \(x^{a(l-1)}\) via a first-order Taylor approximation:
\begin{equation}
J^{e*}(x^a) \approx J^{e*}\!\big(x^{a(l-1)}\big)
+ \left.\frac{\mathrm{d} J^{e*}}{\mathrm{d} x^a}\right|_{x^{a(l-1)}}\!(x^a - x^{a(l-1)}).
\label{eq:linearize}
\end{equation}
Using the Karush-Kuhn-Tucker (KKT) optimality conditions of the ego’s constrained problem (\ref{eq:ego_opt}),  the sensitivity of \(J^{e*}\) with respect to the attacker’s decision can be approximated around the current iterate as:
\begin{equation}
\left.\frac{\mathrm{d} J^{e*}}{\mathrm{d} x^a}\right|_{x^{a(l-1)}} 
\approx -\mu^{e(l)} 
\left.\frac{\partial \gamma_e}{\partial x^a}\right|_{(x^{a(l-1)},x^{e(l)})},
\label{eq:sensitivity}
\end{equation}
where \(\mu^{e(l)}\!\ge\!0\) is the KKT multiplier for ego’s active collision constraint \(\gamma_e(x^a,x^e)\le0\).  
Substituting Eqs.~(\ref{eq:linearize})-(\ref{eq:sensitivity}) yields the attacker’s sensitivity-enhanced update:
\begin{equation}
\begin{aligned}
\hat{x}_0^a \approx\; &\arg\max_{x^a}\; 
J^a(x^a) 
+ \alpha \mu^{e(l)} 
\left.\frac{\partial \gamma_e}{\partial x^a}\right|_{(x^{a(l-1)},x^{e(l)})}
x^a
 \\
\text{s.t.}\quad 
&\text{(same constraints as (\ref{eq:attacker_opt})).}
\end{aligned}
\label{eq:attacker_seibr}
\end{equation}

When the ego’s avoidance constraint becomes active (i.e. \(\mu^{e(l)}>0\)), the attacker acquires a directional incentive proportional to $\partial \gamma_e/{\partial x^a}$, which encourages attacker motions that decrease the signed-distance in the ego--attacker interaction space. 
Intuitively, this steers the attacker toward maneuvers that effectively reduce the ego’s feasible maneuvering set and thus elicit a reactive avoidance response from the ego, while the KL-based trust-region preserves distributional plausibility under the pretrained model. Because the collision-avoidance constraint only activates within the ego’s responsibility region, this design indirectly ensures the generated adversarial behaviors correspond to plausible and meaningful interactions rather than irrelevant collisions (e.g., passive rear-endings). To raise the probability of entering this active regime in practice, we bias attacker route initialization during the Warmup phase toward spatial corridors that intersect the ego’s responsibility region. Once the collision-avoidance constraint is activated, the first-order sensitivity term governs the attacker’s subsequent refinements, yielding targeted yet distributionally consistent adversarial behaviors. Detailed sensitivity derivations are provided in the Appendix.

\section{Experiments}
\label{sec:experiment}

\subsection{Setup and Protocol}

We employ the Nexus model~\cite{zhou2025decoupled}, a diffusion-based driving scene generator pretrained on the nuPlan~\cite{caesar2021nuplan} dataset, and apply our inference procedure directly on its outputs without any additional training or fine-tuning, named Nexus-$\mathrm{\Omega}$. We compare our method with recently available and faithfully reproduced diffusion-based scene generation approaches trained on the nuPlan dataset under multiple settings. Details of our experimental settings and baselines are provided in the Appendix.

\subsection{Comparison to State of the Art}
In this section, we assess the method’s realism and physical plausibility, generalizability to unseen environments, and goal-conditioned controllability.

\paragraph{Free exploration evaluation}
The model generates multiple plausible futures conditioned on the history states in this setting. 
As shown in Table~\ref{tab:comparison}, our approach, denoted as Nexus-$\mathrm{\Omega}$, delivers consistent gains in both realism and physical plausibility over prior diffusion-based generators. 
On nuPlan, it substantially reduces failure cases, cutting scene-level collisions by \textbf{25.76} points and off-road events by \textbf{29.74} points, while improving kinematic feasibility by \textbf{3.3} points. These gains result in a \textbf{72.27\%} valid rate, nearly doubling Nexus' overall performance.
When evaluated zero-shot on the unseen Waymo~\cite{WaymoMotion} dataset, Nexus-$\mathrm{\Omega}$ maintains its advantage, achieving a valid rate of \textbf{81.95\%}, \textbf{+20.24} points higher than Nexus, demonstrating strong generalizability.
To further assess the generality of OMEGA and its effect on distributional realism, we apply it to both Nexus and VBD~\cite{huang2026versatile} and evaluate them on the Waymo Open Sim Agents \textit{val} set, with Nexus evaluated zero-shot. As shown in Table~\ref{tab:wosac}, OMEGA improves both generators. The larger gain for Nexus highlights the greater benefits of guidance for interaction and map-based realism under zero-shot shift.

\begin{table*}[!t]
\centering
\caption{%
    \textbf{Free-exploration scene generation results on nuPlan and zero-shot evaluation on Waymo.} 
    Oracle refers to original logs in the dataset.
    N-Dist. (nearest-agent distance distribution JSD, $10^{-3}$), 
    L-Dev. (lateral deviation distribution JSD, $10^{-3}$), 
    A-Dev. (angular deviation distribution JSD, $10^{-3}$), 
    Spd. (speed distribution JSD, $10^{-3}$), 
    P-Ag./P-Sc. (per-agent/per-scene rate), 
    and overall Scene Valid Rate (no collision/off-road/kinematic violation). 
    $\downarrow$ indicates lower is better; $\uparrow$ indicates higher is better.
}
\label{tab:comparison}
\resizebox{\textwidth}{!}{
\begin{tabular}{c|l|cccc|cc|cc|cc|>{\columncolor{Gray}}c}
\toprule
& & \multicolumn{4}{c|}{\textbf{Distributional JSD}} & \multicolumn{2}{c|}{\textbf{Col.(\%)}} & \multicolumn{2}{c|}{\textbf{Off-road(\%)}} & \multicolumn{2}{c|}{\textbf{K.Feas.(\%)}} & \textbf{Valid(\%)} \\
\multirow{-2}{*}{\textbf{Dataset}} & \multirow{-2}{*}{\textbf{Method}} & N-Dist. $\downarrow$ & L-Dev. $\downarrow$ & A-Dev. $\downarrow$ & Spd. $\downarrow$ & P-Ag. $\downarrow$ & P-Sc. $\downarrow$ & P-Ag. $\downarrow$ & P-Sc. $\downarrow$ & P-Ag. $\uparrow$ & P-Sc. $\uparrow$ & P-Sc. $\uparrow$ \\
\midrule
\multicolumn{1}{l|}{} &
\textcolor{black!60}{Oracle} &
\textcolor{black!60}{0.000} &
\textcolor{black!60}{0.000} &
\textcolor{black!60}{0.000} &
\textcolor{black!60}{0.000} &
\textcolor{black!60}{0.38} &
\textcolor{black!60}{13.62} &
\textcolor{black!60}{0.00} &
\textcolor{black!60}{0.00} &
\textcolor{black!60}{97.40} &
\textcolor{black!60}{81.22} &
\textcolor{black!60}{71.67} \\
\multicolumn{1}{l|}{} & D. Policy~\cite{chi2025diffusion} & 0.701 & 0.056 & 0.423 & 0.369 & 1.19& 36.58& 4.37 & 39.47 & 98.46 & 87.18 & 34.49 \\
\multicolumn{1}{l|}{} & SceneD.~\cite{jiang2024scenediffuser} & 0.933 & 0.046 & 0.395 & \textbf{0.234} & 1.19& 37.80& 4.84 & 41.68 & 98.48 & 86.96 & 33.27 \\
\multicolumn{1}{l|}{} & Nexus~\cite{zhou2025decoupled} & 1.162 & 0.041 & 0.461 & 1.018 & 1.41& 43.12& 4.70 & 40.13 & 98.72 & 88.63 & 32.35 \\
\multicolumn{1}{l|}{\multirow{-5}{*}{nuPlan}} & Nexus-$\mathrm{\Omega}$ & \textbf{0.203} & \textbf{0.025} & \textbf{0.081} & 0.333 & \textbf{0.19}& \textbf{17.36} & \textbf{0.91} & \textbf{10.39} & \textbf{99.19} & \textbf{91.91} & \textbf{72.27} \\
\midrule
& \textcolor{black!60}{Oracle}
& \textcolor{black!60}{0.000}
& \textcolor{black!60}{0.000}
& \textcolor{black!60}{0.000}
& \textcolor{black!60}{0.000}
& \textcolor{black!60}{0.22}
& \textcolor{black!60}{7.47}
& \textcolor{black!60}{0.00}
& \textcolor{black!60}{0.00}
& \textcolor{black!60}{99.77}
& \textcolor{black!60}{96.33}
& \textcolor{black!60}{89.33} \\
& D. Policy~\cite{chi2025diffusion} & 0.034 & 0.026 & 0.081 & 0.229 & 0.35& 19.23& 1.04 & 11.35 & 98.86 & 86.74 & 64.55 \\
& SceneD.~\cite{jiang2024scenediffuser} & 0.077 & 0.022 & 0.148 & 0.249 & 0.39& 22.94 & 1.22 & 12.18 & 98.81 & 86.57 & 61.01 \\
& Nexus~\cite{zhou2025decoupled} & 0.081 & 0.022 & 0.154 & 0.251 & 0.38 & 22.65 & 1.23 & 12.30 & 98.76 & 86.08 & 61.71 \\
\multirow{-5}{*}{Waymo} & Nexus-$\mathrm{\Omega}$ & \textbf{0.019} & \textbf{0.015} & \textbf{0.050} & \textbf{0.198} & \textbf{0.18} & \textbf{11.43} & \textbf{0.37} & \textbf{4.32} & \textbf{99.40} & \textbf{92.65} & \textbf{81.95} \\
\bottomrule
\end{tabular}
}
\end{table*}
\begin{table}[h]
\centering
\caption{\textbf{Distributional realism under the Waymo Open Sim Agents evaluation protocol.} We compare Nexus and VBD with their respective $\mathrm{\Omega}$-enhanced variants.}
\label{tab:wosac}
\resizebox{\linewidth}{!}{%
\begin{tabular}{l|cccc|ccc|cc|c}
\toprule
\multirow{2}{*}{\textbf{Method}}
& \multicolumn{4}{c|}{\textbf{Kinematics} $\uparrow$}
& \multicolumn{3}{c|}{\textbf{Interaction} $\uparrow$}
& \multicolumn{2}{c|}{\textbf{Map-Based} $\uparrow$}
& \multirow{2}{*}{\makecell{\textbf{Realism}\\\textbf{Meta} $\uparrow$}} \\
& Lin. Spd.
& Lin. Acc.
& Ang. Spd.
& Ang. Acc.
& Obj. Dist.
& \ \ \ Col.\ \ \ 
& \ \ \ TTC\ \ \ \ 
& Road Dist.
& \ Offroad\ \ \ 
& \\
\midrule
Nexus
& 0.253
& 0.377
& 0.324
& 0.611
& 0.290
& 0.763
& 0.833
& 0.573
& 0.683
& 0.609 \\
Nexus-$\mathrm{\Omega}$
& 0.246
& 0.321
& 0.329
& 0.598
& 0.302
& 0.884
& 0.835
& 0.609
& 0.919
& 0.700 \\
\midrule
VBD
& 0.353
& 0.332
& 0.437
& 0.563
& 0.341
& 0.943
& 0.820
& 0.631
& 0.895
& 0.723 \\
VBD-$\mathrm{\Omega}$
& 0.352
& 0.353
& 0.441
& 0.567
& 0.358
& 0.951
& 0.823
& 0.640
& 0.898
& 0.730 \\
\bottomrule
\end{tabular}%
}
\end{table}

\paragraph{Goal-conditioned controllability}
In this setting, we specify an explicit goal point for a target agent at its final timestep and adjust the inpainting mask $\tilde{m}$ accordingly. The model is expected to generate a feasible scene completion that retains overall realism under the specified inpainting condition.
Unlike prior approaches that often enforce goal reaching through abrupt trajectory deformation, leading to large lane deviation, off-road or dynamically infeasible motions, Nexus-$\mathrm{\Omega}$ integrates the goal constraint naturally during the denoising process, allowing smooth adherence to the target without compromising feasibility. 
As shown in Table~\ref{tab:controllability}, OMEGA reduces the scene-level collision rate (\textbf{-39.00}) and the off-road rate (\textbf{-44.00}), and boosts kinematic feasibility (\textbf{+36.00}) over Nexus. This improvement results in an increase of \textbf{+69.00} points, nearly an order-of-magnitude gain over the baselines. 
Overall, OMEGA enables precise goal controllability while maintaining natural multi-agent coordination and physical plausibility.

\begin{table}[t]
\centering
\caption{%
\textbf{Comparison on goal-conditioned controllability.} 
Each method is evaluated by setting multiple user-defined goal points for the target vehicle to test controllability. }
\label{tab:controllability}
\resizebox{\linewidth}{!}{
\begin{tabular}{l|cccc|cc|cc|cc|>{\columncolor{Gray}}c}
\toprule
                                  & \multicolumn{4}{c|}{\textbf{Distributional JSD}} & \multicolumn{2}{c|}{\textbf{Col.(\%)}} & \multicolumn{2}{c|}{\textbf{Off-road(\%)}} & \multicolumn{2}{c|}{\textbf{K.Feas.(\%)}} & \textbf{Valid(\%)} \\
\multirow{-2}{*}{\textbf{Method}} & N-Dist. $\downarrow$                  & L-Dev. $\downarrow$                   & A-Dev. $\downarrow$                   & Spd. $\downarrow$                                             & P-Ag. $\downarrow$                    & P-Sc. $\downarrow$                    & P-Ag. $\downarrow$                    & P-Sc. $\downarrow$                   & P-Ag. $\uparrow$                      & P-Sc. $\uparrow$                      & P-Sc. $\uparrow$                      \\ \midrule
{D. Policy~\cite{chi2025diffusion}}  & {  0.382}          & {  0.706}          & {  0.047}          & {  73.245}                                 & {  0.81}          & {  40.00}          & {  3.71}          & {  46.00}         & {  96.32}          & {  58.00}          & {  20.00}          \\
{SceneD.~\cite{jiang2024scenediffuser}}    & {  0.526}          & {  1.405}          & {  0.033}          & {  75.818}                                 & {  0.88}          & {  46.00}          & {  4.21}          & {  40.00}         & {  96.63}          & {  59.00}          & {  17.00}          \\
{Nexus~\cite{zhou2025decoupled}}      & {  0.594}          & {  1.618}          & {  0.037}          & {  76.187}                                 & {  1.20}          & {  51.00}          & {  5.22}          & {  51.00}         & {  96.44}          & {  55.00}          & {  11.00}          \\
Nexus-$\mathrm{\Omega}$                       & {  \textbf{0.222}} & {  \textbf{0.244}} & {  \textbf{0.012}} &   \textbf{5.151} & {  \textbf{0.10}} & {  \textbf{12.00}} & {  \textbf{0.44}} & {  \textbf{7.00}} & {  \textbf{99.37}} & {  \textbf{91.00}} & {  \textbf{80.00}} \\ \bottomrule
\end{tabular}
}
\end{table}

\subsection{Adversarial Scenario Generation}

\begin{figure*}[!t]
\caption{%
    \textbf{Adversarial scene generation results.} \textbf{(left)} Comparison between oracle (GT), Nexus variants, including: Nexus finetuned on adversarial scenarios (Nexus-FT); Nexus with goal-attacking condition (Nexus-GC); Nexus-CTG with goal-attacking cost (Nexus-CTG$_\mathrm{Adv}$). 
    Ego Risk: mean time to collision per frame (TTC, $s$), percentage of TTC $<1,2,3\,s$. 
    Intensity of Ego Motion: mean accleration (Acc, $m^2/s$), jerk ($m^3/s$). 
    Ego Non-responsible Collision (Ego NC, \%).
    \textbf{(right)} Nexus-$\mathrm{\Omega}_\mathrm{Adv}$ produces more samples with short TTC and low speed, indicating the generation of riskier scenarios.
}
\label{tab:adv_generation}
\begin{minipage}[t]{0.60\textwidth}
\vspace{0pt}
\centering
\resizebox{\linewidth}{!}{
    \begin{tabular}{l|cccc|cc|ccc}
    \toprule
    & \multicolumn{4}{c|}{\textbf{Ego Risk}} & \multicolumn{2}{c|}{\textbf{Ego Motion}} & \textbf{Ego NC} & \textbf{Off-road} & \textbf{K. Feas.} \\
    \multirow{-2}{*}{\textbf{Method}} & TTC $\downarrow$ & \textless 1s $\uparrow$ & \textless 2s $\uparrow$ & \textless 3s $\uparrow$ & Acc. $\uparrow$ & Jrk. $\uparrow$ & P-Sc. $\downarrow$ & P-Sc. $\downarrow$ & P-Sc. $\uparrow$ \\
    \midrule
    \rowcolor{gray!15}
    Oracle & 4.904 & 0.328 & 0.949 & 2.212 & 0.360 & 0.307 & 0.97 & 0.00 & 81.22 \\
    \midrule
    Nexus-FT & 4.695 & 3.239 & 5.453 & 7.214 & 0.355 & 0.477 & 9.91 & 39.77 & 87.08 \\
    Nexus-GC & 4.631 & 3.462 & 6.179 & 8.677 & 0.631 & 1.384 & 10.98 & 45.14 & 45.16 \\
    Nexus-CTG$_\mathrm{Adv}$ & 4.575 & 3.541 & 6.254 & 8.747 & 0.794 & \textbf{2.145} & 7.12 & 26.77 & 85.64 \\
    Nexus-$\mathrm{\Omega}_\mathrm{Adv}$ & \textbf{4.516} & \textbf{3.450} & \textbf{7.721} & \textbf{11.599} & \textbf{0.859} & 1.972 & \textbf{5.42} & \textbf{15.82} & \textbf{88.98} \\
    \bottomrule
    \end{tabular}
}
\end{minipage}
\hfill
\begin{minipage}[t]{0.38\textwidth}
\vspace{-3pt}
    \centering
    \includegraphics[width=0.99\linewidth]{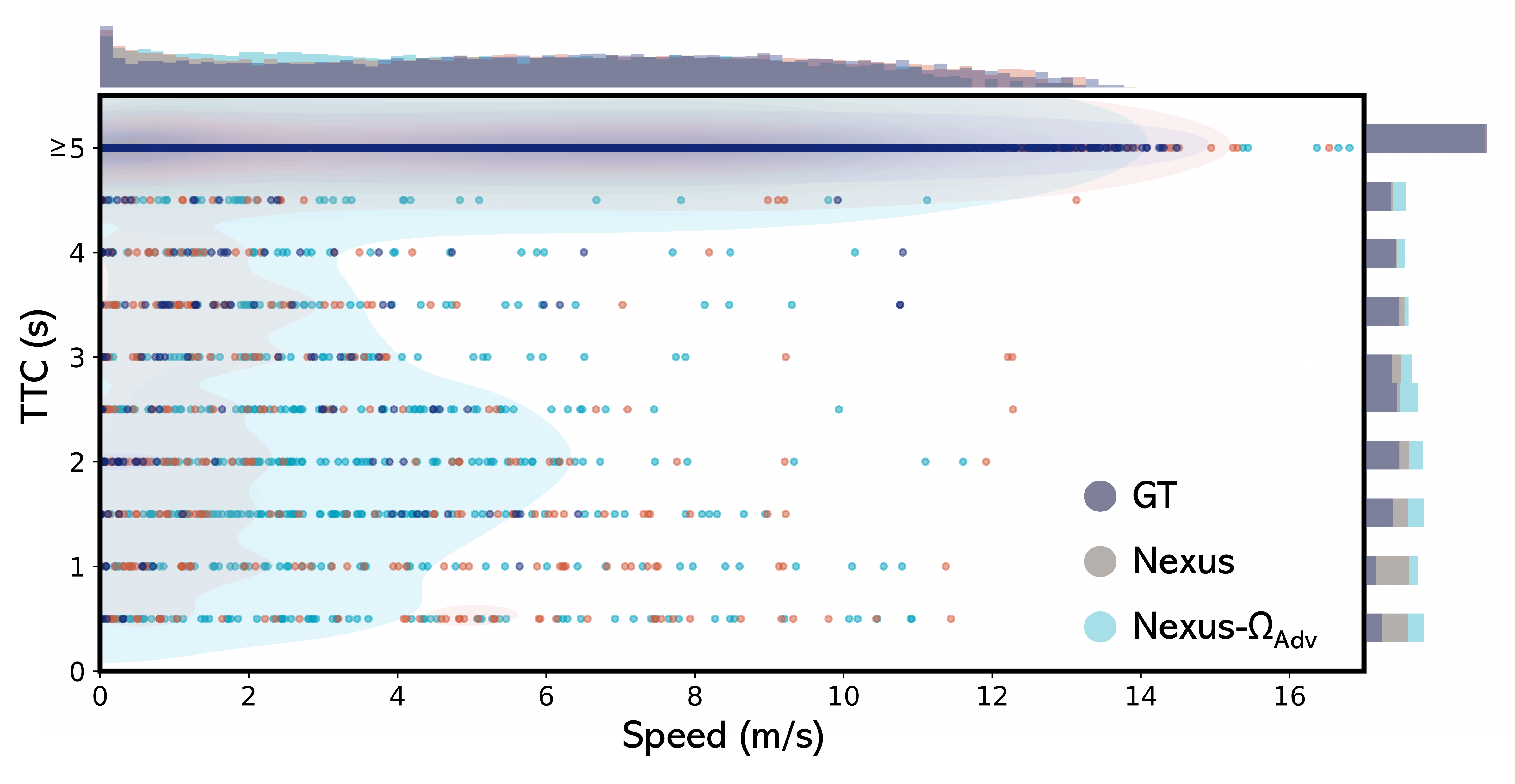}
\end{minipage}
\vspace{-6pt}
\end{figure*}

\paragraph{Generic adversarial generation} To characterize the strength and realism of the generated adversarial long-tail interactions, we compare Nexus-$\mathrm{\Omega}_\mathrm{Adv}$ with multiple baselines. As shown in the left and right panels of \cref{tab:adv_generation}, our guidance strategies produce substantially more interactive scenes than the GT distribution and other Nexus baselines. This is evidenced by consistently lower mean TTC and higher proportions of ego TTC below 1–3 seconds, indicating increased risk exposure for the ego vehicle. 
Correspondingly, ego trajectory statistics, including mean acceleration and jerk increase, suggest that generated scenarios trigger stronger evasive or aggressive responses. 
Importantly, despite the increased attack intensity, off-road rates and kinematic feasibility remain acceptable (lower off-road, higher kinematic feasibility), and the Ego NC metric confirms that the generated collisions are more meaningful--reflecting realistic, ego-responsible interactions that provide greater training and evaluation value.
Together, these results demonstrate that our method substantially increases scenario aggressiveness while preserving physical plausibility.

\begin{wraptable}[12]{r}{0.49\columnwidth}
\centering
\vspace{-10pt}
\caption{\textbf{Planner-in-the-loop adversarial generation results.} Three ego planners, log replay (Replay), constant velocity (CV), and IDM, are evaluated on Log and Nexus-$\mathrm{\Omega}_\mathrm{Adv}$. Scene-level minimum TTC and collision rates are reported, with collisions split into ego-at-fault (A.F.) and non-at-fault (N.A.F.).}
\label{tab:planner_loop_adv}
\setlength{\tabcolsep}{3pt}
\small
\resizebox{\linewidth}{!}{%
\begin{tabular}{l|l|cccc|cc}
\toprule
\multirow{2}{*}{Planner} & \multirow{2}{*}{Scenarios} & \multicolumn{4}{c|}{Scene Min TTC (s)} & \multicolumn{2}{c}{Scene Collision (\%)} \\
 &  & \textless{}1s$\uparrow$ & \textless{}2s$\uparrow$ & \textless{}3s$\uparrow$ & Avg$\downarrow$ & A.F.$\uparrow$ & N.A.F.$\downarrow$ \\
\midrule
\multirow{2}{*}{Replay} & Log & 0.07 & 1.85 & 4.72 & 4.97 & 0.02 & \textbf{0.97} \\
 & Nexus-$\mathrm{\Omega}_\mathrm{Adv}$ & \textbf{21.31} & \textbf{25.52} & \textbf{29.14} & \textbf{4.64} & \textbf{19.50} & 4.47 \\
\midrule
\multirow{2}{*}{CV} & Log & 14.82 & 16.88 & 19.06 & 4.79 & 8.00 & 10.35 \\
 & Nexus-$\mathrm{\Omega}_\mathrm{Adv}$ & \textbf{23.17} & \textbf{26.98} & \textbf{29.87} &\textbf{ 4.61} & \textbf{17.25} & \textbf{6.34} \\
\midrule
\multirow{2}{*}{IDM} & Log & 6.51 & 8.83 & 14.04 & 4.89 & 1.34 & 8.98 \\
 & Nexus-$\mathrm{\Omega}_\mathrm{Adv}$ & \textbf{18.55} & \textbf{22.34} & \textbf{25.66} & \textbf{4.64} & \textbf{15.62} & \textbf{5.37} \\
\bottomrule
\end{tabular}%
}
\end{wraptable}
\paragraph{Planner-specific adversarial generation} By injecting the ego planned trajectory into the game as a weak-noise condition, our framework enables planner-specific adversarial generation.
As shown in Table~\ref{tab:planner_loop_adv}, Nexus-$\mathrm{\Omega}_\mathrm{Adv}$ consistently increases risk exposure across different planners by shifting scene minimum TTC toward shorter values and raising ego-at-fault collision rates.
Meanwhile, the non-at-fault collision rate remains low, suggesting planner-relevant breakdowns rather than uninformative passive collisions.

\subsection{Further Analysis}

\begin{table}[!t]
\centering
\caption{
    \textbf{Comparison of different guidance strategies} and \textbf{ablation study}. Results under the \textit{Free Exploration} setting.
}
\label{tab:guidance_comparison}
\resizebox{\linewidth}{!}{
\begin{tabular}{l|cccc|cc|cc|cc|>{\columncolor{Gray}}c }
\toprule
                                     & \multicolumn{4}{c|}{\textbf{Distributional JSD}} & \multicolumn{2}{c|}{\textbf{Col.(\%)}} & \multicolumn{2}{c|}{\textbf{Off-road(\%)}} & \multicolumn{2}{c|}{\textbf{K.Feas.(\%)}} & \textbf{Valid(\%)} \\
\multirow{-2}{*}{\textbf{Method}}    & N-Dist. $\downarrow$ & L-Dev. $\downarrow$ & A-Dev. $\downarrow$ & Spd. $\downarrow$  & P-Ag. $\downarrow$ & P-Sc. $\downarrow$      & P-Ag. $\downarrow$ & P-Sc. $\downarrow$     & P-Ag. $\uparrow$ & P-Sc. $\uparrow$                 & P-Sc. $\uparrow$      \\
\midrule
Nexus~\cite{zhou2025decoupled} & 1.162          & 0.041          & 0.461          & 1.018          & 1.41          & 43.12          & 4.70          & 40.13          & 98.72          & 88.63          & 32.35          \\
Nexus-GHC~\cite{jiang2024scenediffuser}                      & 2.368          & 0.041          & 0.694          & 3.222          & 2.23          & 49.05          & 1.58          & 11.57          & 92.50          & 50.29          & 26.06          \\
Nexus-CTG~\cite{zhong2022guided}                      & 0.392          & 0.031          & 0.225          & 0.713          & \underline{0.44}          & \underline{26.02}          & 0.92          & \underline{9.97}  & 97.13          & 81.72          & 50.41          \\
Nexus-DPS~\cite{zheng2025diffusion}                      & 0.421          & 0.028          & 0.214          & \underline{0.677}          & 0.51          & 27.25          & 0.95          & 11.34          & 97.56          & 83.55          & 47.19          \\
\midrule
w/o Rolling-Zero & \underline{0.251} & \underline{0.028} & \underline{0.108} & 2.168 & 1.19 & 34.35 & \textbf{0.57} & \textbf{6.64} & \textbf{99.69} & \textbf{96.72} & \underline{60.49} \\
w/o Warm-up       & 23.180 & 0.178 & 113.874 & 184.708 & 8.18 & 76.90 & 45.01 & 98.95 & 3.15 & 0.19 & 0.00 \\
w/o Guid. Obj.   & 2.241 & 0.094 & 0.248 & 1.285 & 1.37 & 41.38 & 4.66 & 39.72 & 98.49 & 85.69 & 33.47 \\
w/o Phase Align.   & 0.705 & 0.046 & 1.132 & 3.061 & 0.68 & 31.54 & 1.20 & 13.03 & 97.07 & 75.73 & 52.05 \\
w/o KL cons.       & 2.544 & 0.060 & 6.773 & 53.086 & 3.15 & 63.40 & 5.43 & 46.38 & 89.97 & 51.95 & 18.21 \\
\midrule
Nexus-$\mathrm{\Omega}$                 & \textbf{0.203} & \textbf{0.025} & \textbf{0.081} & \textbf{0.333} & \textbf{0.19} & \textbf{17.36} & \underline{0.91} & 10.39 & \underline{99.19} & \underline{91.91} & \textbf{72.27} \\
\bottomrule
\end{tabular}
}
\end{table}

\textbf{Comparison with other guidance methods.}
As shown in \cref{tab:guidance_comparison}, Nexus-$\mathrm{\Omega}$ achieves the best overall performance among the projection-style guidance method GHC~\cite{jiang2024scenediffuser}, the conditional score approximation method CTG~\cite{zhong2022guided}, and the prior clean-space guidance method DPS~\cite{zheng2025diffusion}. It substantially improves realism, while reducing collision and off-road rates and improving kinematic feasibility. These consistent gains translate into a markedly higher overall valid rate compared with prior guidance approaches. The results suggest that optimization-guided diffusion more effectively balances realism with constraint satisfaction, yielding physically consistent scenes.

\paragraph{Ablation study}
As shown in \cref{tab:guidance_comparison}, removing the Rolling-Zero phase leads to smoother but less responsive trajectories: the model performs well on-road but exhibits higher collision rates. Using only the Rolling-Zero phase (without Warm-up) results in severe cumulative autoregressive errors and degraded performance across all metrics. Retaining the two-phase noise schedule without guidance objectives yields only marginal gains over the base model, indicating that noise scheduling alone is insufficient. Applying guidance without phase-specific alignment retains only part of the scene-validity gain while requiring \textbf{$7.85\times$} the inference time of the full OMEGA configuration, increasing from $4.96{\pm}3.92,\mathrm{s}$ to $38.94{\pm}39.79,\mathrm{s}$, highlighting the benefits of phase-aligned guidance for both generation quality and computational efficiency. Omitting the KL divergence constraint drastically reduces distributional realism and overall stability. The full OMEGA configuration achieves the best balance among realism, smoothness, and safety, yielding the highest valid rate.

\section{Conclusion}
We present OMEGA, a training-free, optimization-guided framework to enhance the fidelity, interactivity, and controllability of diffusion-based driving scene generation. A two-stage noise schedule further enhances interaction realism by combining global trajectory refinement with local frame-level responsiveness. OMEGA$_\mathrm{Adv}$ then produces realistic and challenging long-tail scenarios. A limitation of our current implementation is its runtime: optimizing an 8-second scene can take around 5 seconds. Parallelizing our per-agent optimization can reduce this latency, enabling the integration of OMEGA into closed-loop simulation \mbox{frameworks. We will leave it as future work}.

\section*{Acknowledgments}

This work is supported by the China Scholarship Council and the National Natural Science Foundation of China (Grant Nos. 52172378 and 52302489). This work is in part supported by the JC STEM Lab of Autonomous Intelligent Systems funded by The Hong Kong Jockey Club Charities Trust.

We sincerely thank Yuxuan Hu, Yuhang Lu, and Haochen Liu for their valuable assistance, and we extend our appreciation to the members of OpenDriveLab for their continuous support throughout this study.

\bibliographystyle{splncs04}
\bibliography{main}

\clearpage

\setcounter{tocdepth}{2}

\clearpage
\appendix
\renewcommand*{\theHsection}{appendix.\Alph{section}}
\renewcommand*{\theHsubsection}{\theHsection.\arabic{subsection}}

%
\providecommand{\authcount}[1]{}
\startcontents[appendix]
{
\hypersetup{linkcolor=black}
\begin{center}
{\large\bfseries Optimization-Guided Diffusion for Interactive Scene Generation}\\[2.7ex]
{\normalfont\fontsize{11.5}{13.5}\selectfont Supplementary Material}
\end{center}
\vspace{1.5em}
\printcontents[appendix]{}{1}{}
}

\clearpage

\section{Theoretical Foundations}
\subsection{Derivation of Optimization-Guided Diffusion}
\label{app:omega_derivation}

This section provides a detailed derivation of the
optimization-guided reverse diffusion operator used in
OMEGA.  
We make explicit how the KL-constrained re-anchoring
of Gaussian reverse kernels induces the Euclidean
trust-region formulation in Eq.~\eqref{eq:euclid-form}, and how the noise
schedule shapes the resulting optimization geometry.

\vspace{3pt}
\paragraph{Re-anchored reverse kernel}
For a fixed reverse step \(t\), denote the current noisy
state by \(x_t\), and let the denoising network produce
a clean estimate \(\tilde{x}_0\).
The standard DDPM reverse kernel can be written as
\begin{equation}
Q_t(\tilde{x}_0)
=
\mathcal{N}\!\bigl(
A_t \tilde{x}_0 + C_t x_t,\;
\sigma_t^2 I
\bigr),
\label{eq:q_kernel_app}
\end{equation}
where the scalar coefficients \(A_t, C_t\) and the variance
\(\sigma_t^2\) are determined by the diffusion schedule.

OMEGA introduces a refined anchor \(x\) and defines a
family of re-anchored kernels
\begin{equation}
P_t(x)
=
\mathcal{N}\!\bigl(
A_t x + C_t x_t,\;
\sigma_t^2 I
\bigr).
\label{eq:p_kernel_app}
\end{equation}
The two kernels share the same isotropic covariance and
differ only in their mean, which allows for a closed-form
expression of their KL divergence.

\vspace{3pt}
\paragraph{Closed-form KL Divergence}
For two Gaussians
\(P = \mathcal{N}(\mu_P,\Sigma)\),
\(Q = \mathcal{N}(\mu_Q,\Sigma)\)
with identical covariance \(\Sigma\),
the KL divergence simplifies to
\begin{equation}
\mathrm{KL}(P \,\|\, Q)
=
\frac{1}{2}
(\mu_P - \mu_Q)^{\top}
\Sigma^{-1}
(\mu_P - \mu_Q).
\label{eq:kl_general}
\end{equation}
Setting \(\Sigma = \sigma_t^2 I\) and using
Eqs.~\eqref{eq:q_kernel_app}–\eqref{eq:p_kernel_app},
we have
\begin{align}
\mu_P - \mu_Q
&=
\bigl(A_t x + C_t x_t\bigr)
-
\bigl(A_t \tilde{x}_0 + C_t x_t\bigr)
\nonumber \\
&=
A_t (x - \tilde{x}_0).
\label{eq:mean_diff}
\end{align}
Substituting Eq.~\eqref{eq:mean_diff} into
Eq.~\eqref{eq:kl_general} gives
\begin{align}
\mathrm{KL}\!\left(
P_t(x)\,\|\,Q_t(\tilde{x}_0)
\right)
&=
\frac{1}{2}
\bigl(A_t (x - \tilde{x}_0)\bigr)^{\top}
\bigl(\sigma_t^2 I\bigr)^{-1}
A_t (x - \tilde{x}_0)
\nonumber \\
&=
\frac{A_t^2}{2\sigma_t^2}
\,
\|x - \tilde{x}_0\|_2^2.
\label{eq:kl_expanded_app}
\end{align}
Thus, up to a scalar factor, the KL divergence is exactly
the squared Euclidean distance between \(x\) and
\(\tilde{x}_0\).

\vspace{3pt}
\paragraph{Equivalent Euclidean formulation of the KL-regularized update}
From the closed-form expression in Eq.~\eqref{eq:kl_expanded_app},
the KL trust-region constraint
\(\mathrm{KL}(P_t(x)\|Q_t(\tilde{x}_0))\le\kappa_t\)
is equivalent to the Euclidean ball
\begin{equation}
\|x-\tilde{x}_0\|_2
\;\le\;
\frac{\sqrt{2\kappa_t}\,\sigma_t}{|A_t|}.
\label{eq:euclidean_tr_app}
\end{equation}

To incorporate structural and behavioral preferences,
the guidance functional \(R(x)\) in Eq.~\eqref{eq:omega_opt}
is typically instantiated
through an augmented objective widely used in constrained
optimization, where equality constraints
\(h(x)=0\) and inequality constraints \(g(x)\le0\) are absorbed
into the objective via quadratic penalties and logarithmic
barrier terms:
\begin{equation}
R(x)
=
r(x)
-
\frac{\rho_h}{2}\|h(x)\|_2^2
-
\rho_g \sum_i \phi\!\bigl(g_i(x)\bigr),
\label{eq:augmented_R_app}
\end{equation}
with \(\phi\) a smooth barrier.  
This decomposition clarifies the structure encoded in
\(R(x)\) and makes explicit the feasibility conditions
governing the refinement variable \(x\).

Substituting Eq.~\eqref{eq:kl_expanded_app} into the
KL-regularized objective
\begin{equation}
\max_x\;
\lambda_t R(x)
-
\mathrm{KL}\!\left(P_t(x)\|Q_t(\tilde{x}_0)\right),
\label{eq:kl_obj_app}
\end{equation}
and reinstating the explicit representation of
\(r(x)\), \(h(x)\), and \(g(x)\) from
Eq.~\eqref{eq:augmented_R_app}, we obtain the
equivalent Euclidean trust-region program
\begin{align}
\hat{x}_0
&=
\arg\max_x
\Bigl[
\lambda_t r(x)
-
\frac{A_t^2}{2\sigma_t^2}
\|x-\tilde{x}_0\|_2^2
\Bigr]
\nonumber \\
&\quad\text{s.t.}\;
\|x-\tilde{x}_0\|_2
\le
\frac{\sqrt{2\kappa_t}\,\sigma_t}{|A_t|},\nonumber\\
&\qquad
h(x)=0,\;
g(x)\le0.
\label{eq:euclidean_prog_app}
\end{align}

This establishes the Euclidean form of the KL-regularized
refinement step: a quadratic stabilizer centered at
\(\tilde{x}_0\) coupled with a noise-scaled trust region
whose radius is proportional to \(\sigma_t/|A_t|\), jointly
governing the allowable deviation from the pretrained
reverse transition at each denoising stage.

\paragraph{Noise-scaled geometry of the refinement step}
Eq.~\eqref{eq:euclidean_prog_app} shows that the refinement
operator is governed by two noise-dependent geometric
components:  
a quadratic stabilizer centered at \(\tilde{x}_0\) and an
explicit Euclidean trust region whose radius is also coupled to
the noise level.

The quadratic term  
\(
\frac{A_t^{2}}{2\sigma_t^{2}}\|x-\tilde{x}_0\|_2^{2}
\)
induces an effective curvature
\(
A_t^{2}/\sigma_t^{2}
\),
which grows as \(\sigma_t\) decreases.  
This progressively sharpens the local optimization landscape,
forcing the anchor update to remain increasingly concentrated
around the model prediction.  
Conversely, the trust-region constraint  
\(
\|x-\tilde{x}_0\|_2
\le
\frac{\sqrt{2\kappa_t}\,\sigma_t}{|A_t|}
\)
scales linearly in \(\sigma_t/|A_t|\), shrinking the allowable
step size as the reverse process proceeds.

These two effects act in concert:  
at early high-noise stages, the stabilizer is relatively flat and
the trust region is wide, allowing substantial adjustments that
can reshape scene-level semantics and multi-agent interaction
patterns;  
at later low-noise stages, both the curvature amplification and
the contracted trust region confine the refinement to a narrow
vicinity of \(\tilde{x}_0\), encouraging localized corrections while limiting deviation from the learned generative manifold.

Together, this establishes OMEGA as a noise-adaptive
trust-region operator whose optimization landscape evolves
smoothly with the diffusion variance, enabling coarse global
adjustments early in denoising and fine-grained refinements as
the process converges.
This intrinsic noise dependence allows the refinement to adapt over the denoising process with fixed hyperparameters. Accordingly, we set \(\lambda_t \equiv 1.0\) and \(\kappa_t \equiv 0.2\) for all experiments.

\subsection{Derivation of the Sensitivity Term}
\label{app:sensitivity}

The sensitivity term used in SE-IBR quantifies how the ego’s optimal refinement value responds to perturbations in the attacker’s refinement. It provides the first-order correction necessary to approximate the Nash equilibrium of the underlying noncooperative game in our distributional formulation. We derive this term below.

\vspace{3pt}
\paragraph{From the coupled game to a reduced best-response objective}
The coupled problems in Eqs.~\eqref{eq:ego_opt}--\eqref{eq:attacker_opt}
define a noncooperative game in the space of
refined anchors \((x^e,x^a)\).
A Nash equilibrium \((x^{e*},x^{a*})\) is a pair such that
no player can improve its own objective by unilaterally
changing its decision within its feasible set:
\begin{equation}
\begin{aligned}
x^{e*} &\in 
\arg\max_{x^e \in \mathcal{X}^e(x^{a*})} J^e(x^e), \\
x^{a*} &\in 
\arg\max_{x^a \in \mathcal{X}^a(x^{e*})}
\bigl[J^a(x^a) - \alpha J^e(x^e)\bigr].
\end{aligned}
\label{eq:app_nash}
\end{equation}
where \(\mathcal{X}^e(x^a)\) and \(\mathcal{X}^a(x^e)\) are the
feasible sets induced by the trust-region constraint,
the general equality and inequality constraints
\(h(\cdot)\) and \(g(\cdot)\), and the coupled safety constraints
\(\gamma_e,\gamma_a\). Although \(\gamma_e\) and \(\gamma_a\) are structurally inequality constraints and may be viewed as specific instances within \(g(\cdot)\), they depend explicitly on both players' refinement
variables and therefore transmit interaction effects between the two optimization problems.

For each fixed attacker decision \(x^a\), the ego’s best
response is
\[
x^{e*}(x^a)
\in
\arg\max_{x^e \in \mathcal{X}^e(x^a)} J^e(x^e),
\]
and the associated optimal value is
\(J^{e*}(x^a) \triangleq J^e(x^{e*}(x^a))\).
Substituting this relation into the attacker’s problem rewrites the game from the attacker’s perspective as
\begin{equation}
x^{a*}
\in
\arg\max_{x^a \in \mathcal{X}^a(x^{e*}(x^a))}
\Bigl[J^a(x^a) - \alpha J^{e*}(x^a)\Bigr].
\label{eq:app_reduced_game}
\end{equation}
Thus the attacker optimizes its own adversarial reward \(J^a\)
while penalizing the ego’s best-response performance through
the term \(J^{e*}(x^a)\), which implicitly captures how the attacker’s decision modifies the ego’s best-response.

Since \(J^{e*}\) is not available in closed form, SE-IBR performs
first-order linearization at the current attacker iterate
\(x^{a(l-1)}\):
\begin{equation}
J^{e*}(x^a)
\approx
J^{e*}\!\bigl(x^{a(l-1)}\bigr)
+
\left.
\frac{\mathrm{d} J^{e*}}{\mathrm{d} x^a}
\right|_{x^{a(l-1)}}
\!\bigl(x^a - x^{a(l-1)}\bigr),
\label{eq:app_linearize}
\end{equation}
which motivates the derivation of the sensitivity term
\(\mathrm{d}J^{e*}/\mathrm{d}x^a\).

\paragraph{Sensitivity of the ego’s optimal value}
To evaluate how the ego’s optimal value changes with the
attacker’s decision, we treat \(x^a\) as the parameter of the
ego-side refinement problem Eq.~\eqref{eq:ego_opt}. 
Within this parametrization, the safety constraint
\(\gamma_e(x^e,x^a)\) provides the explicit functional coupling
between the two players and therefore dominates the first-order
sensitivity.  We consider a neighborhood of the iterate
\(x^{a(l-1)}\) in which the coupled safety constraint is active,
\(\gamma_e(x^{e(l)},x^{a(l-1)}) = 0\), and the active set of all
other constraints remains unchanged.  Under this standard
assumption in parametric sensitivity analysis of nonlinear
programs~\cite{spica2020real}, the active inequality
\(\gamma_e(x^e,x^a)\le 0\) can locally be treated as an equality
constraint for the purpose of computing derivatives.

Let \(x^{e(l)} = x^{e*}(x^{a(l-1)})\) denote the optimal ego
refinement at iteration \(l\), and define the value function
\(J^{e*}(x^a) = J^e(x^{e*}(x^a))\).  
By the chain rule,
\begin{equation}
\left.
\frac{\mathrm{d}J^{e*}}{\mathrm{d}x^a}
\right|_{x^{a(l-1)}}
=
\left.
\frac{\partial J^e}{\partial x^e}
\right|_{x^{e(l)}}
\cdot
\left.
\frac{\mathrm{d}x^{e*}}{\mathrm{d}x^a}
\right|_{x^{a(l-1)}}.
\label{eq:app_chainrule}
\end{equation}

Since \(x^{e(l)}\) solves the constrained program, it satisfies
the KKT conditions of Eq.~\eqref{eq:ego_opt}.
Let \(\mu^{e(l)}\ge 0\) be the multiplier associated with the
active interaction constraint \(\gamma_e(x^e,x^a)\le 0\), and let
\(\lambda^e,\nu^e\) collect the multipliers corresponding to all
other equality and inequality constraints that do not explicitly
depend on \(x^a\).  The stationarity condition at
\((x^{e(l)},x^{a(l-1)})\) can be written as
\begin{equation}
\begin{aligned}
\left.\frac{\partial J^e}{\partial x^e}\right|_{x^{e(l)}}
&\;-\;
\mu^{e(l)}
  \left.\frac{\partial \gamma_e}{\partial x^e}
  \right|_{(x^{e(l)},x^{a(l-1)})} \\
&\;-\;
\lambda^{e\top}
  \left.\frac{\partial h}{\partial x^e}\right|_{x^{e(l)}}
\;-\;
\nu^{e\top}
  \left.\frac{\partial g}{\partial x^e}\right|_{x^{e(l)}}
= 0.
\end{aligned}
\label{eq:app_kkt_stationarity}
\end{equation}

We now right-multiply \eqref{eq:app_kkt_stationarity} by
\(\mathrm{d}x^{e*}/\mathrm{d}x^a\).  
For any constraint that does not explicitly depend on \(x^a\),
its value at the optimal solution satisfies, e.g.
\(h(x^{e*}(x^a)) = 0\), so taking the total derivative with
respect to \(x^a\) gives
\(\frac{\partial h}{\partial x^e}\,\frac{\mathrm{d}x^{e*}}{\mathrm{d}x^a}=0\),
and analogously for the active components of \(g(\cdot)\).
Hence the corresponding Jacobian terms vanish after
right-multiplication, and we obtain
\begin{equation}
\begin{aligned}
\left.
\frac{\partial J^e}{\partial x^e}
\right|_{x^{e(l)}}
\cdot
\left.
\frac{\mathrm{d}x^{e*}}{\mathrm{d}x^a}
\right|_{x^{a(l-1)}}
&=
\mu^{e(l)}
\left.
\frac{\partial \gamma_e}{\partial x^e}
\right|_{(x^{e(l)},x^{a(l-1)})}
\\[2pt]
&\hspace{1.3cm}\cdot\;
\left.
\frac{\mathrm{d}x^{e*}}{\mathrm{d}x^a}
\right|_{x^{a(l-1)}}.
\end{aligned}
\label{eq:app_proj_stationarity}
\end{equation}
By \eqref{eq:app_chainrule}, the left-hand side is exactly
\(\mathrm{d}J^{e*}/\mathrm{d}x^a\).

On the other hand, in the neighborhood where
\(\gamma_e\) is active we have
\(\gamma_e(x^{e*}(x^a),x^a)=0\), so differentiating with respect
to \(x^a\) yields
\begin{equation}
\left.
\frac{\partial \gamma_e}{\partial x^e}
\right|_{(x^{e(l)},x^{a(l-1)})}
\cdot
\left.
\frac{\mathrm{d}x^{e*}}{\mathrm{d}x^a}
\right|_{x^{a(l-1)}}
+
\left.
\frac{\partial \gamma_e}{\partial x^a}
\right|_{(x^{e(l)},x^{a(l-1)})}
= 0,
\label{eq:app_constraint_derivative}
\end{equation}
which implies
\[
\left.
\frac{\partial \gamma_e}{\partial x^e}
\right|_{(x^{e(l)},x^{a(l-1)})}
\cdot
\left.
\frac{\mathrm{d}x^{e*}}{\mathrm{d}x^a}
\right|_{x^{a(l-1)}}
=
-
\left.
\frac{\partial \gamma_e}{\partial x^a}
\right|_{(x^{e(l)},x^{a(l-1)})}.
\]

Substituting this identity into
\eqref{eq:app_proj_stationarity} finally gives
\begin{equation}
\left.
\frac{\mathrm{d}J^{e*}}{\mathrm{d}x^a}
\right|_{x^{a(l-1)}}
=
-
\mu^{e(l)}
\left.
\frac{\partial \gamma_e}{\partial x^a}
\right|_{(x^{e(l)},x^{a(l-1)})},
\label{eq:app_final_sensitivity}
\end{equation}
which is the sensitivity term used in SE-IBR.

\section{Toy Experiment}

Following the two-dimensional toy setup used to visualize generative sampling dynamics in~\cite{holderrieth2025introduction}, we design a controlled experiment to examine how OMEGA influences the geometry of the learned generative distribution. Unlike high-dimensional driving scenes, where such dynamics are difficult to visualize, this simplified setting enables a clear analysis of the sampling behavior under guidance. In particular, the experiment investigates two questions: (i) whether optimization-based re-anchoring can steer the sampling distribution toward directions favored by the guidance objective, and (ii) whether the KL-bounded refinement can regulate the guidance while limiting drift from the learned distribution. To this end, we compare three sampling regimes: (a) unguided diffusion, (b) OMEGA re-anchoring with the objective but without the KL constraint, and (c) the full OMEGA formulation, and analyze how each regime affects the denoising dynamics and resulting sample distribution.
\cref{fig:toy_exm} summarizes the outcomes across these settings.

\subsection{Experiment Setup}
We consider a controlled two-dimensional generation task in which the diffusion sampling process is initialized from a standard Gaussian prior, while the target distribution consists of five well-separated Gaussian components arranged in a predefined geometric configuration. Each component represents a distinct mode, collectively forming a structured multimodal distribution. We train a lightweight MLP-based diffusion model to generate samples from this target distribution, providing a simple and interpretable setting for analyzing how guidance alters the learned distribution.

To study how guidance affects the learned distribution, we introduce a guidance point, shown as the yellow star in \cref{fig:toy_exm}, and define a reward function \(R(x)\) that encourages samples to approach it. The guidance point is deliberately placed outside the high-density regions of all five Gaussian components, but closer to one component than to the others, thereby imposing a clear directional preference on the guided sampling process. This setup creates an explicit trade-off: the reward \(R(x)\) pulls samples toward the guidance point, whereas the diffusion model favors samples that remain consistent with the original target distribution. Such a controlled conflict allows us to examine whether a guidance mechanism can accommodate the external preference without collapsing or significantly distorting the underlying modes.

\subsection{Results Across Sampling Regimes}
\paragraph{Unguided Diffusion}
In the baseline configuration (\cref{fig:toy_exm}(a)), sampling follows the standard reverse diffusion process without external guidance. The model successfully recovers the target five-mode distribution, with sample trajectories gradually denoising toward the five Gaussian components and the final samples forming five clearly separated modes. This result indicates that the diffusion model has accurately learned the essential structure of the target distribution.

\paragraph{Optimization-Based Re-Anchoring without KL Constraint}
In the second setting (\cref{fig:toy_exm}~(b)), we activate the guidance point and apply OMEGA’s optimization-based re-anchoring using only the objective term $R(x)$, without the KL constraint. Under this configuration, re-anchoring incorporates the guidance objective into each reverse diffusion update, introducing a directional preference toward the guidance point. As a result, the intermediate sampling distribution at each denoising step is progressively biased toward regions favored by $R(x)$, and the sampled trajectories exhibit the corresponding directional tendency.
However, without the KL constraint, the magnitude of these optimization updates is not regulated, and their effects accumulate across denoising steps. As sampling progresses, the resulting trajectories increasingly drift away from the Gaussian modes of the target distribution, and the geometry of the guided sampling dynamics becomes progressively distorted. By the final denoising steps (e.g., \(t=0.01\) in \cref{fig:toy_exm}~(b)), many samples have moved away from the Gaussian components and into low-density regions between the learned modes, reflecting a weakening of the original mode structure and an increasing risk of mode collapse during denoising.

\paragraph{Re-Anchoring with KL Constraint}
In the final setting (\cref{fig:toy_exm}~(c)), we apply OMEGA with both the objective term $R(x)$ and the KL constraint. In this configuration, the KL constraint bounds the distributional shift introduced by re-anchoring at each denoising step. As illustrated in \cref{fig:toy_exm}~(c), re-anchoring continues to steer the sampling distribution toward the guidance direction while largely preserving the overall structure of the target distribution: the five Gaussian modes remain well separated throughout denoising, without the pronounced distortion observed in the unconstrained case. Rather than collapsing onto the guidance point, samples concentrate in a nearby region that achieves higher reward while remaining close to the learned modes, indicating that the KL constraint regulates the re-anchoring updates and limits excessive distributional drift.

\subsection{Interpretation and Takeaways}

This toy experiment provides an intuitive illustration of the mechanism underlying OMEGA’s guidance strategy. Optimization-based re-anchoring incorporates the guidance signal into the reverse diffusion updates, allowing the intermediate sampling distribution to gradually shift toward regions favored by the objective  $R(x)$ .
The KL constraint moderates this process by regulating the magnitude of each distributional update. By bounding the shift of the sampling distribution at every denoising step, it limits excessive drift from the learned distribution and mitigates mode collapse during intermediate denoising stages.

Taken together, these two components play complementary roles, enabling OMEGA to guide the sampling process in a stable manner while avoiding severe distortion of the learned distribution.

\section{Implementation Details}

This section summarizes the components used by OMEGA during inference.  
We first describe the pretrained Nexus model that serves as the generative backbone, then introduce the guidance objectives used for optimization-guided re-anchoring, and finally detail the two-phase noise schedule that integrates global denoising with local interaction adaptation.

\subsection{Base Generative Model}
\label{app:nexus}

We apply OMEGA on top of the pretrained {Nexus} scene generator~\cite{zhou2025decoupled}, a diffusion-based multi-agent inpainting model trained on large-scale real-world driving logs from the nuPlan dataset~\cite{caesar2021nuplan}. We adopt Nexus as the base generator because it is a publicly available diffusion-based generative framework with released checkpoints, enabling fair and reproducible comparisons. We use the publicly released Nexus checkpoints without additional training. The resulting system, referred to as {Nexus-$\mathrm{\Omega}$}, performs all refinement purely at inference time via optimization-guided reverse denoising.

Nexus represents each multi-agent traffic scene as a 
spatiotemporal tensor of agent states, including position, 
heading, velocity, and physical size.  Conditional generation is 
performed under flexible inpainting masks that specify arbitrary 
subsets of known elements, such as observed history frames, agent 
identities, or goal states.  At every diffusion step, Nexus 
predicts the noise residual associated with these agent tokens 
given the scene context, map information, and the inpainting 
structure; the clean estimate is obtained using the standard diffusion reconstruction.

The model was pretrained on over
{1200 hours of real-world nuPlan driving logs}, consisting 
of 10-second scenarios sampled at 2\,Hz. The training 
procedure employs a {tri-axial noise sampling} strategy that 
assigns independent noise levels across agents, timesteps, and 
feature dimensions.  This exposes the model to a wide spectrum of 
partially denoised configurations, including heterogeneous and 
nonuniform noise patterns distributed over the sequence.  
Consequently, Nexus learns robust denoising behaviors over diverse 
noise assignments, a property that is naturally compatible with 
the per-frame and time-varying noise configurations introduced in 
our inference procedure.  

These extensive training regimes provide Nexus with strong priors 
over multi-agent behavior, including lane-following patterns, 
interaction dynamics, and map-conditioned scene structure.  
OMEGA leverages these priors by refining each predicted clean 
sample through a constraint-aware re-anchoring procedure, thereby 
improving physical and behavioral fidelity without altering the backbone model.

\subsection{Guidance Objectives Used in OMEGA}
\label{app:guidance}

Following the Euclidean formulation in Eq.~\eqref{eq:euclid-form}, the
refinement step in OMEGA is formulated as an optimization problem with explicit equality and inequality constraints. We express the guidance model as
\[
R(x) \;\equiv\; r(x) \quad\text{with}\quad h(x)=0,\;\; g(x)\le 0,
\]
where \(r(x)\) encodes trajectory-level preference signals, while
\(h(x)\) and \(g(x)\) capture structural constraints among agent states and feasible behavioral constraints.  Together, they impose a set of regularization constraints on the refined multi-agent trajectories.

In our implementation, these constraints and objectives reflect five principal aspects of scene structure:  
(i) \textit{kinematic consistency} enforced through bicycle-model auxiliary dynamics;  
(ii) \textit{traffic-rule compliance}, including heading alignment
and on-road adherence;  
(iii) \textit{multi-agent safety}, modeled through geometric
separation constraints;  
(iv) \textit{state-consistency constraints} for initial, terminal,
or intermediate anchoring; and  
(v) \textit{motion smoothness} regulating the control-related auxiliary variables.  
All components are differentiable and combine with
the KL trust-region constraint, supporting stable refinement within
the reverse denoising process. We describe each component in detail below.

\paragraph{Kinematic consistency}
We introduce virtual accelerations \(a_\tau\) and steering angles
\(\delta_\tau\), together with an estimated wheelbase \(L\) obtained
from the predicted vehicle length.  
These quantities act as auxiliary control variables governing the temporal evolution of the refined trajectories; they are not predicted by the diffusion model but are introduced solely to impose a physically interpretable structure on the refinement process.

Let the refined trajectory at physical timestep \(\tau\) be represented by the
state
\[
s_\tau = \bigl(p^x_\tau,\, p^y_\tau,\, \psi_\tau,\, v_\tau\bigr),
\]
where \((p^x_\tau,p^y_\tau)\) denotes position, \(\psi_\tau\) the heading,
and \(v_\tau\) the speed.  
The local motion is governed by the bicycle model (applied for each timestep \(\tau\))
\[
\begin{aligned}
\dot{p}^x_\tau &= v_\tau\cos\psi_\tau, &
\dot{p}^y_\tau &= v_\tau\sin\psi_\tau, \\
\dot{\psi}_\tau &= \tfrac{v_\tau}{L}\tan\delta_\tau, &
\dot{v}_\tau &= a_\tau.
\end{aligned}
\]

Discretizing with step \(\Delta \tau\) yields the kinematic residual sequence  
\[
h_{\mathrm{kin}}(x)
=
\bigl\{
\xi^{\mathrm{kin}}_\tau(x)
\bigr\}_{\tau=0}^{\mathcal{T}-1},
\]
where each residual vector is
\[
\xi^{\mathrm{kin}}_\tau(x)
=
\begin{bmatrix}
p^x_{\tau+1}-p^x_\tau-\Delta \tau\, v_\tau\cos\psi_\tau \\
p^y_{\tau+1}-p^y_\tau-\Delta \tau\, v_\tau\sin\psi_\tau \\
\psi_{\tau+1}-\psi_\tau-\Delta \tau\, \tfrac{v_\tau}{L}\tan\delta_\tau \\
v_{\tau+1}-v_\tau-\Delta \tau\, a_\tau
\end{bmatrix}.
\]

The equality constraint \(h_{\mathrm{kin}}(s)=0\) therefore
requires the refined trajectories to satisfy the auxiliary control variables and therefore follow the
underlying bicycle-model dynamics.

\paragraph{Traffic-rule compliance}
To evaluate traffic-rule compliance, we first enumerate all
feasible forward route candidates for each agent.  
Each candidate route \(\mathcal{C}^{(m)}=\{c^{(m)}_k\}\) is a
sequence of reference centerline points obtained via depth-first search over lane-successor graphs.  
At each refinement time step \(\tau\), the refined position
\((p^x_\tau,p^y_\tau)\) is associated with the closest reference point
over all candidates:
\[
c_\tau^\ast
=
\arg\min_{c \in \cup_m \mathcal{C}^{(m)}}
\bigl\|(p^x_\tau,p^y_\tau)-c\bigr\|_2.
\]
This provides a locally consistent reference position and heading
for every time step.

\textit{Heading alignment.}
Let \(\psi_\tau^{\mathrm{ref}}\) denote the heading associated with
the reference point \(c_\tau^\ast\).
For high-speed motion, the deviation \(|\psi_\tau - \psi_\tau^{\mathrm{ref}}|\) is required to
stay within a tolerance \(\psi_{\max}\), as large deviations correspond to unrealistic or unsafe maneuvers
\[
g_{\mathrm{head}}(x)
=
\bigl\{\,\xi^{\mathrm{head}}_\tau(x)\,\bigr\}_{\tau=0}^{\mathcal{T}},
\]
where
\[
\xi^{\mathrm{head}}_\tau(x)
=
|\psi_\tau - \psi_\tau^{\mathrm{ref}}| - \psi_{\max}.
\]

\textit{On-road driving.}
Let \(d_\tau^{\mathrm{lat}}\) denote the lateral distance from
\(x_\tau\) to the drivable region corresponding to \(c_\tau^\ast\).
To ensure that moving agents remain on the road surface within a tolerated margin \(b_{\max}\), the inequality constraint is
\[
g_{\mathrm{road}}(x)
=
\bigl\{\,\xi^{\mathrm{road}}_\tau(x)\,\bigr\}_{\tau=0}^{\mathcal{T}},
\]
where
\[
\xi^{\mathrm{road}}_\tau(x)
=
d_\tau^{\mathrm{lat}} - b_{\max}.
\]

\paragraph{Multi-agent safety}
Inter-agent collision avoidance is modeled by approximating each
agent \(i\) using two safety circles 
\(\{o_{i,\tau}^{(1)},\, o_{i,\tau}^{(2)}\}\) at each time step \(\tau\),
each with radii \(r_o^{(i)}\).
For any pair of distinct agents \(i\neq j\), define the
center-to-center distances
\[
d_{ij,\tau}^{(k,k')}
=
\bigl\|o_{i,\tau}^{(k)} - o_{j,\tau}^{(k')}\bigr\|_2,
\qquad
r_{\mathrm{sum}}^{(i,j)} = r_o^{(i)} + r_o^{(j)},
\]
for \(k,k'\in\{1,2\}\).
Collision-free motion requires that every pair of circles maintain a separation of at least their combined radii.  
This yields the time-indexed collection of inequality residuals
\[
g_{\mathrm{safe}}(x)
=
\bigl\{
\xi^{\mathrm{safe}}_{ij\tau k k'}(x)
\bigr\}_{i\neq j,\;\tau,\;k,k'},
\]
where
\[
\xi^{\mathrm{safe}}_{ij,\tau,k,k'}(x)
=
r_{\mathrm{sum}} - d_{ij,\tau}^{(k,k')}.
\]
In the attacker–ego scenario, the corresponding safety constraint used in the sensitivity analysis (see \cref{app:sensitivity}) is of the same form, and is written as \(\gamma_e(x^e,x^a)\) when applied to the ego–attacker circle pairs.

\paragraph{State-consistency constraints}
When specific states are prescribed at selected timesteps, e.g., initial or terminal states, they are enforced via equality constraints of the form
\[
h_{\mathrm{state}}(x)
=
\bigl\{
\xi^{\mathrm{state}}_\tau(x)
\bigr\}_{\tau \in \mathcal{I}},
\qquad
\xi^{\mathrm{state}}_\tau(x)
=
s_\tau - \bar{s}_\tau,
\]
where \(\bar{s}_\tau\) denotes the desired state, and \(\mathcal{I}\) is the set of timesteps where constraints are applied. This formulation encompasses common boundary specifications,
including \(s_0 = s_{\mathrm{init}}\) and
\(s_{\mathcal{T}} = s_{\mathrm{goal}}\), and accommodates intermediate anchoring constraints when required by the scenario or task.  These constraints ensure that the refined trajectory remains consistent with externally imposed prescribed state conditions.

\paragraph{Motion smoothness}
To encourage physically consistent motion, the auxiliary control
variables \(a_\tau\) (longitudinal acceleration) and
\(\delta_\tau\) (steering angle) are regularized through a
smoothness objective that penalizes their magnitudes as well as their temporal variations
\[
\begin{aligned}
r_{\mathrm{smooth}}(x)
=
\sum_{\tau=0}^{\mathcal{T}-1}
\bigl(
&\;w_a\, a_\tau^2
+ w_{a\mathrm{d}}\, (a_{\tau+1}-a_\tau)^2 \\
&+ w_\delta\, \delta_\tau^2
+ w_{\delta\mathrm{d}}\, (\delta_{\tau+1}-\delta_\tau)^2
\bigr).
\end{aligned}
\]
where \(w_a, w_{a\mathrm{d}}, w_\delta, w_{\delta\mathrm{d}}>0\)
weight the trade-off between control-effort magnitude and temporal smoothness.  
This regularization encourages smooth temporal evolution in acceleration
and steering, preventing abrupt changes that would be physically implausible or unrealistic.

\paragraph{Unified constrained formulation}
Combining the above components, the structured guidance objective consists of the smoothness objective
\[
r(x)=r_{\mathrm{smooth}}(x),
\]
together with equality constraints
\[
h(x)=\{\,h_{\mathrm{kin}}(x),\; h_{\mathrm{state}}(x)\,\}=0,
\]
and inequality constraints
\[
g(x)=
\{\,g_{\mathrm{head}}(x),\;
g_{\mathrm{road}}(x),\;
g_{\mathrm{safe}}(x)\,\}
\;\le\; 0.
\]

\paragraph{Penalty-based realization}
To ensure numerical stability and maintain compatibility with
the KL-based trust-region constraint in Eq.~\eqref{eq:euclid-form},
the constrained formulation is implemented through a combination of explicit constraints and differentiable penalty terms.
Equality conditions are enforced via a quadratic penalty of the form
\[
w_h\,\|h(x)\|_2^2,
\]
while inequality conditions that are softly enforced are handled through
\[
w_g\, [g(x)]_+^{\,2},
\qquad
[z]_+ = \max(0,z).
\]
This hybrid treatment avoids hard projection operations that may conflict with the adaptive trust-region bound, and preserves stability throughout the refinement process throughout the reverse denoising
process.

\subsection{Phase-Aligned Guidance Schedule}
\label{app:two_phase_schedule}

OMEGA integrates its optimization-guided refinement with a two-phase denoising schedule that aligns guidance objectives with different stages of scene formation. Rather than applying a uniform denoising strategy across the entire horizon, the schedule separates macro-level motion layout formation from fine-grained interaction refinement. This phase-aligned design improves both global motion plausibility and local interaction responsiveness. The complete two-phase procedure is summarized in Algorithm~\ref{alg:omega_two_phase}.

\paragraph{Warm-up phase with agent-wise structural guidance}
The Warm-up phase begins from the maximally corrupted scene state $x_T$ and performs reverse denoising along the sequence $\{T, \dots, t_{\mathrm{low}}\}$. At each diffusion level $t$, the model predicts a clean estimate $\tilde{x}_0$, which is refined by the KL-bounded re-anchoring procedure of OMEGA to obtain $\hat{x}_0$. The refined anchor then defines the reverse transition $x_{t-1} \sim \mathcal{N}(A_t \hat{x}_0 + C_t x_t,\,
\sigma_t^2 I)$.

During the Warm-up phase, the guidance objective $R_{\mathrm{warm}}$ includes only agent-wise structural components, such as kinematic consistency, traffic-rule compliance, state-consistency constraints, and motion smoothness, all enforced jointly across the entire future horizon. Since all timesteps share the same diffusion level, each reverse update operates on the full trajectory, encouraging macro-level spatial organization and stable long-horizon motion trends.
The schedule intentionally stops at an intermediate noise level $t_{\mathrm{low}}$, preserving controlled residual uncertainty and providing a well-conditioned initialization for the subsequent interaction refinement stage.

\paragraph{Rolling-Zero phase with inter-agent interaction guidance}
Starting from the Warm-up output $x_{t_{\mathrm{low}}}$, the Rolling-Zero phase progressively removes the remaining uncertainty through a time-indexed denoising schedule. For each timestep $\tau = 1{:}\mathcal{T}$, a noise-assignment vector $\mathbf{t}^{(\tau)} \in \{0,\, t_{\mathrm{low}}\}^{\mathcal{T}}$ is constructed, where past and current frames $\{1{:}\tau\}$ are assigned noise level $0$, while future frames $\{\tau{+}1{:}\mathcal{T}\}$ retain the residual noise $t_{\mathrm{low}}$. This schedule freezes previously refined states, isolates the update to the current frame, and preserves a controlled amount of uncertainty for the frames that will be refined later.

Given the partially denoised scene state $x^{\mathrm{roll}}$ and the per-frame noise configuration $\mathbf{t}^{(\tau)}$, the model predicts a clean estimate $\tilde{x}_0^{(\tau)}$ consistent with this per-frame noise assignment.  
OMEGA then performs KL-bounded re-anchoring using a guidance objective $R_{\mathrm{roll}}^{(\tau)}$ that augments the structural objectives with inter-agent interaction terms. These interaction terms are evaluated using the full scene state $x^{\mathrm{roll}}$, providing up-to-date reference trajectories that allow each agent to assess potential conflicts with others. This design decouples the per-agent refinements at frame $\tau$, permits parallel optimization across agents, and ensures that the updates remain aligned with the most recent global scene
configuration.

Only the components corresponding to frame $\tau$ are updated under the reverse transition defined by $\mathbf{t}^{(\tau)}$, and the resulting scene state initializes the next step $\tau{+}1$. By advancing this refinement across the horizon, Rolling-Zero gradually resolves inter-agent interactions while preserving the globally coherent motion structure established during Warm-up.

\begin{algorithm}[t]
\caption{Phase-Aligned Guidance Schedule with OMEGA}
\label{alg:omega_two_phase}
\begin{algorithmic}[1]
\Require Incomplete scene tensor $\bar{x}$, validity mask $m$, inpainting mask $\tilde{m}$, context inputs $c$
\Ensure Refined future states $\hat{s}_{1:\mathcal{T}}$

\State Initialize the noisy scene representation $x_T \sim \mathcal{N}(0,I)$

\Statex
\Statex \textbf{Warm-up phase with agent-wise structural guidance}
\For{$t = T, T-1, \dots, t_{\mathrm{low}}$}
    \State $\tilde{x}_0 \leftarrow \textsc{ModelPredictX0}(x_t,\, t,\; \bar{x}, m, \tilde{m}, c)$
    \State $\hat{x}_0 \leftarrow \textsc{OptimizeAnchor}(\tilde{x}_0,\, x_t,\, t,\; R_{\mathrm{warm}})$
    \State $x_{t-1} \sim \mathcal{N}\!\left(A_t \hat{x}_0 + C_t x_t,\; \sigma_t^2 I\right)$
\EndFor
\State $x^{\mathrm{roll}} \leftarrow x_{t_{\mathrm{low}}}$

\Statex
\Statex \textbf{Rolling-Zero phase with inter-agent interaction guidance}
\For{$\tau = 1, \dots, \mathcal{T}$}
    \State Construct interaction context at step $\tau$ from $x^{\mathrm{roll}}$
\Comment{\small for inter-agent safety terms in $R_{\mathrm{roll}}^{(\tau)}$}
    \State $\mathbf{t}^{(\tau)} \leftarrow \textsc{RollingSchedule}(\tau,\; t_{\mathrm{low}},\; \mathcal{T})$
\Comment{\small noise level: $0$ for $1\!:\!\tau$, $t_{\mathrm{low}}$ for $\tau{+}1\!:\!\mathcal{T}$}
    \State $\tilde{x}_0^{(\tau)} \leftarrow \textsc{ModelPredictX0}(x^{\mathrm{roll}},\, \mathbf{t}^{(\tau)},\; \bar{x}, m, \tilde{m}, c)$
    \State $\hat{x}_0^{(\tau)} \leftarrow \textsc{OptimizeAnchor}(\tilde{x}_0^{(\tau)},\, x^{\mathrm{roll}},\, \mathbf{t}^{(\tau)},\; R_{\mathrm{roll}}^{(\tau)})$
    \State $x^{\mathrm{roll}} \sim
    \mathcal{N}\!\big(A_{\mathbf{t}^{(\tau)}} \odot \hat{x}_0^{(\tau)}
    + C_{\mathbf{t}^{(\tau)}} \odot x^{\mathrm{roll}},\;
    \sigma_{\mathbf{t}^{(\tau)}}^2 I\big)$
\EndFor

\State $\hat{s}_{1:\mathcal{T}} \leftarrow \textsc{ExtractStates}(x^{\mathrm{roll}})$
\State \Return $\hat{s}_{1:\mathcal{T}}$
\end{algorithmic}
\end{algorithm}

\section{Experiments}

This section provides implementation and evaluation details
supplementing the results in the main paper.
We describe the protocol used for scene generation and
simulation, define all quantitative metrics, and detail the
experimental setups for each evaluation task.
Our experiments cover three major settings: 
(i) \textit{free-exploration scene generation}, evaluating realism
and physical plausibility on nuPlan with zero-shot transfer to
Waymo;
(ii) \textit{goal-conditioned controllability}, assessing an agent's
ability to reach designated targets under interaction constraints;
(iii) \textit{adversarial scenario generation}, evaluating the synthesis of realistic yet safety-critical attacker–ego interactions, including both generic and planner-specific adversarial settings.


\subsection{Protocols and Metrics}
\label{sec:protocols_metrics}

We evaluate generated scenes along two complementary aspects:
(i) distributional realism with respect to real driving logs, and
(ii) physical and behavioral plausibility of the resulting
multi-agent motion.  
All metrics are computed directly from the generated trajectories
and aggregated over the full evaluation set.

\paragraph{Distributional realism}
Following prior work~\cite{rowe2025scenario,lu2024scenecontrol}, we evaluate how closely the generated scenes match real-world driving logs using the Jensen--Shannon Divergence (JSD):
\begin{equation}
    \mathrm{JSD}(a\|b)
    =
    \frac{1}{2}\mathrm{KL}(a\|c)
    +
    \frac{1}{2}\mathrm{KL}(b\|c),
    \qquad
    c=\frac{1}{2}(a+b),
\end{equation}
where \(a\) and \(b\) denote the empirical distributions of real and generated data, respectively. For each statistic, we aggregate per-timestep values over all valid agents to construct a histogram-based distribution for the real data and each generated sample. We then compute the JSD between each generated distribution and the real-data distribution and report the average across samples. Four statistics are considered:

\begin{itemize}
    \item \textbf{Speed (Spd.)}  
    The instantaneous speed of each valid agent at each timestep.

    \item \textbf{Nearest-agent distance (N-Dist.)}  
    For each agent and timestep, the Euclidean distance to the
    closest other valid agent in the same frame.

    \item \textbf{Lateral deviation (L-Dev.)}  
    The perpendicular distance from an agent's position to the
    nearest lane centerline segment.

    \item \textbf{Angular deviation (A-Dev.)}  
    The absolute orientation difference between the agent’s heading
    and the direction of its nearest centerline segment.
\end{itemize}

\paragraph{Physical and behavioral plausibility}
We assess whether generated trajectories adhere to physical
constraints and exhibit behavior consistent with drivable,
non-colliding motion.  Each metric is
computed at both the agent level and the scene
level.  A scene is considered valid only if all agents in
that scene satisfy the corresponding agent-level criterion.

\begin{itemize}
    \item \textbf{Collision rate.}  
    Agent-level collisions are detected by checking whether an
    agent's bounding box overlaps with any other agent at any
    timestep.  The per-agent collision rate is defined as the
    fraction of agents that collide at least once.  
    A scene is collision-free only if none of its agents collide.

    \item \textbf{Off-road rate.}  
    An agent is marked off-road if its trajectory leaves the
    drivable region at any timestep.  The per-agent off-road rate
    counts the fraction of agents exhibiting at least one off-road
    event.  
    A scene is off-road–free only if every agent remains within
    the drivable region throughout the trajectory.

    \item \textbf{Kinematic feasibility.}  
    For each agent, we derive its temporal kinematic quantities from its predicted trajectory, including velocity, acceleration, jerk, yaw rate, lateral acceleration, and curvature. An agent is considered kinematically feasible only when all of these quantities remain within pre-specified physical limits. The per-agent feasibility rate measures the proportion of agents that satisfy this condition, and a scene is considered kinematically feasible only if every agent in the scene meets the feasibility criterion.
\end{itemize}

\paragraph{Scene Valid Rate}
A scene is counted as valid only if it satisfies all three
conditions above—no collisions, no off-road events, and full
kinematic feasibility for every agent.  We report the
\textit{Scene Valid Rate} as the proportion of scenes that meet
all criteria.

\paragraph{Realism Meta score}
A distribution-based metric used in WOSAC. Given a 1-second history, the model generates 32 stochastic futures whose distribution is compared with the ground-truth distribution across statistics such as speed, collision frequency, and off-road frequency.

\paragraph{Adversarial-specific metrics}
For attacker–ego experiments, we report additional metrics that
characterize the difficulty and risk imposed on the ego vehicle:

\begin{itemize}
    \item \textbf{Time-to-collision (TTC).}  
    TTC is computed by forward-extrapolating agent motions under a
constant-velocity model and detecting the earliest predicted collision with the ego vehicle. TTC values are clipped to a fixed prediction horizon. Lower TTC indicates higher interaction risk.

    \item \textbf{Ego motion intensity.}  
    The ego’s kinematic response is quantified using its mean acceleration and mean jerk computed over all valid timesteps, reflecting the amount of forced maneuvering required to avoid the attacker.

    \item \textbf{Ego non-responsible collision rate.}  
    Using nuPlan’s responsibility rules, collisions are categorized as ego-at-fault or non-responsible. Lower ego non-responsible collision rates indicate fewer irrelevant or passive crashes, while ego-at-fault collisions reflect failures to avoid collisions under adversarial pressure.
\end{itemize}

\subsection{Evaluation Tasks}
\paragraph{Free-exploration scene generation}
In this setting, the model is tasked with generating an
unconstrained 8-second future scene conditioned only on the
historical observations.  
Following the evaluation protocol commonly adopted on nuPlan, we
provide the past 2 seconds of all valid agents and the local
lane-centerline geometry as context, and the generator freely
predicts the multi-agent evolution at 0.5-second intervals.
Agents that are marked invalid at a given timestep are excluded
from both prediction and metric computation.

For evaluation on nuPlan, we randomly sample 4{,}096 scenarios
from the validation split and generate four samples for each
scenario, yielding 16{,}384 future scenes.  
To assess generalization, we apply exactly the same inference
procedure in a zero-shot manner to 4{,}096 scenarios from the
Waymo Motion dataset, again producing four samples per scene.
No retraining or fine-tuning is performed in either case.

We compare OMEGA against the pretrained Nexus generator~\cite{zhou2025decoupled} as well as the re-implementations of Diffusion Policy~\cite{chi2025diffusion}
and SceneDiffuser~\cite{jiang2024scenediffuser}.  
All methods operate under the same conditioning, scene masking,
and evaluation protocol, enabling a controlled comparison of
distributional realism and physical plausibility.

\paragraph{Goal-conditioned controllability}
To assess controllability under structured objectives, we conduct goal-conditioned generation on the nuPlan validation set. We randomly select 25 scenarios and designate one or two vehicles (including the ego) as controlled agents. For each scenario, we define four distinct sets of target goal points, yielding 100 test cases in total.

Goal conditions are imposed through the inpainting formulation,
where the terminal states of controlled agents are specified as
fixed tokens in the input mask. The generator must complete the
remaining future states so that the designated agents reach their assigned goals while maintaining coherent interactions with surrounding traffic.

We compare OMEGA against the pretrained Nexus generator~\cite{zhou2025decoupled} as well as Diffusion Policy~\cite{chi2025diffusion} and SceneDiffuser~\cite{jiang2024scenediffuser}, with all methods
evaluated under identical goal specifications and inpainting
setup. This task measures each method's ability to satisfy long-horizon target constraints without compromising multi-agent realism or physical plausibility.

\paragraph{Generic adversarial generation}
We evaluate adversarial generation under the same free-exploration
setup.  
For each nuPlan scene, we identify a pool of vehicles within a
prescribed distance to the ego and sample one uniformly as the
attacker. All remaining agents
retain their original conditioning.

We compare OMEGA with three Nexus-based adversarial variants.
(i) Nexus-FT is obtained by finetuning the pretrained Nexus model
on Nexus-Data~\cite{zhou2025decoupled}, which contains 540 hours of adversarial driving
scenarios generated using CAT~\cite{zhang2023cat}.
(ii) Nexus-GC applies the goal-attacking strategy of
Nexus~\cite{zhou2025decoupled}, selecting an attacking goal from a heading-aligned sector containing collision-prone future
positions and injecting it into the inpainting mask to steer the
attacker. (iii) Nexus-CTG$_\mathrm{Adv}$~\cite{zhong2022guided} adds a distance-based adversarial cost whose gradient is injected into the reverse-step mean update, encouraging the designated attacker to move closer to the ego and steering the denoising trajectory toward more adversarial interactions.
All methods are evaluated using the adversarial metrics described in \cref{sec:protocols_metrics}, including TTC profiles, ego motion intensity, ego non-responsible collisions, and the standard physical-plausibility checks covering collision rate, off-road violations, and kinematic feasibility. Together, these metrics quantify both the risk level of the generated adversarial interactions and the physical plausibility of the resulting trajectories.

\paragraph{Planner-specific adversarial generation}
Our framework further enables planner-specific adversarial generation targeting downstream planning systems. We therefore conduct a planner-in-the-loop adversarial evaluation where the ego planned trajectory is incorporated into the game formulation as a condition via noise-based partial masking.

We evaluate three ego planners: log replay (Replay), a constant-velocity baseline (CV), and the Intelligent Driver Model (IDM). For each planner, we compare the resulting behaviors on real driving logs and under adversarial scenarios generated by Nexus-$\mathrm{\Omega}_\mathrm{Adv}$.
Evaluation focuses on scene minimum TTC and collision
responsibility statistics, including ego-at-fault and
non-responsible collision rates. These metrics indicate whether
the generated scenarios lead to planner-relevant failures and
elevated risk exposure, rather than incidental or passive
collisions.

\paragraph{Waymo Open Sim Agents evaluation} 
Waymo Open Sim Agents (WOSAC) evaluates scene generation by requiring 32 stochastic future predictions conditioned on 1 second of historical observations at 10 Hz. We evaluate 400 dynamic scenes after screening the reference trajectories for collision and off-road events, interpolate our 2 Hz outputs to match the WOSAC protocol, and run VBD~\cite{huang2026versatile} using its publicly released implementation. While we report WOSAC results for completeness, we do not adopt it as the primary evaluation metric for two reasons.

First, our evaluation protocol is designed to comprehensively assesses scene validity, controllability-conditioned generation, and adversarial scene generation, which extend beyond WOSAC’s scope.

Second, WOSAC defines realism through distributional alignment with human reference trajectories via a weighted meta score, where collision and off-road frequencies constitute high-impact components. 
However, prior analysis \cite{cornelisse2026humanlikeness} indicates that a non-trivial fraction of WOMD scenarios contain collision or off-road events likely attributable to annotation artifacts. In such settings, a strategy that explicitly reduces collision and off-road events may deviate from empirical frequencies and therefore receive a lower meta score despite exhibiting improved safety behavior. For example, \cite{cornelisse2026humanlikeness} shows that a hypothetical strategy identical to human behavior but without collisions or off-road events can obtain a lower meta score of 0.72 than one reproducing the empirical collision frequency which has a meta score of 0.82. 
Since OMEGA explicitly guides generation to suppress collision and off-road events, such structural improvements are not necessarily fully reflected under a purely distribution-consistency metric. 

For these reasons, we treat WOSAC as a complementary benchmark and adopt Scene Valid Rate as our primary metric. It provides a direct measure of structural consistency and physical plausibility in generated scenes, supporting consistent evaluation across nuPlan and Waymo settings.

\subsection{Qualitative Results}

\begin{figure*}[b]
    \centering
    \includegraphics[width=\linewidth]{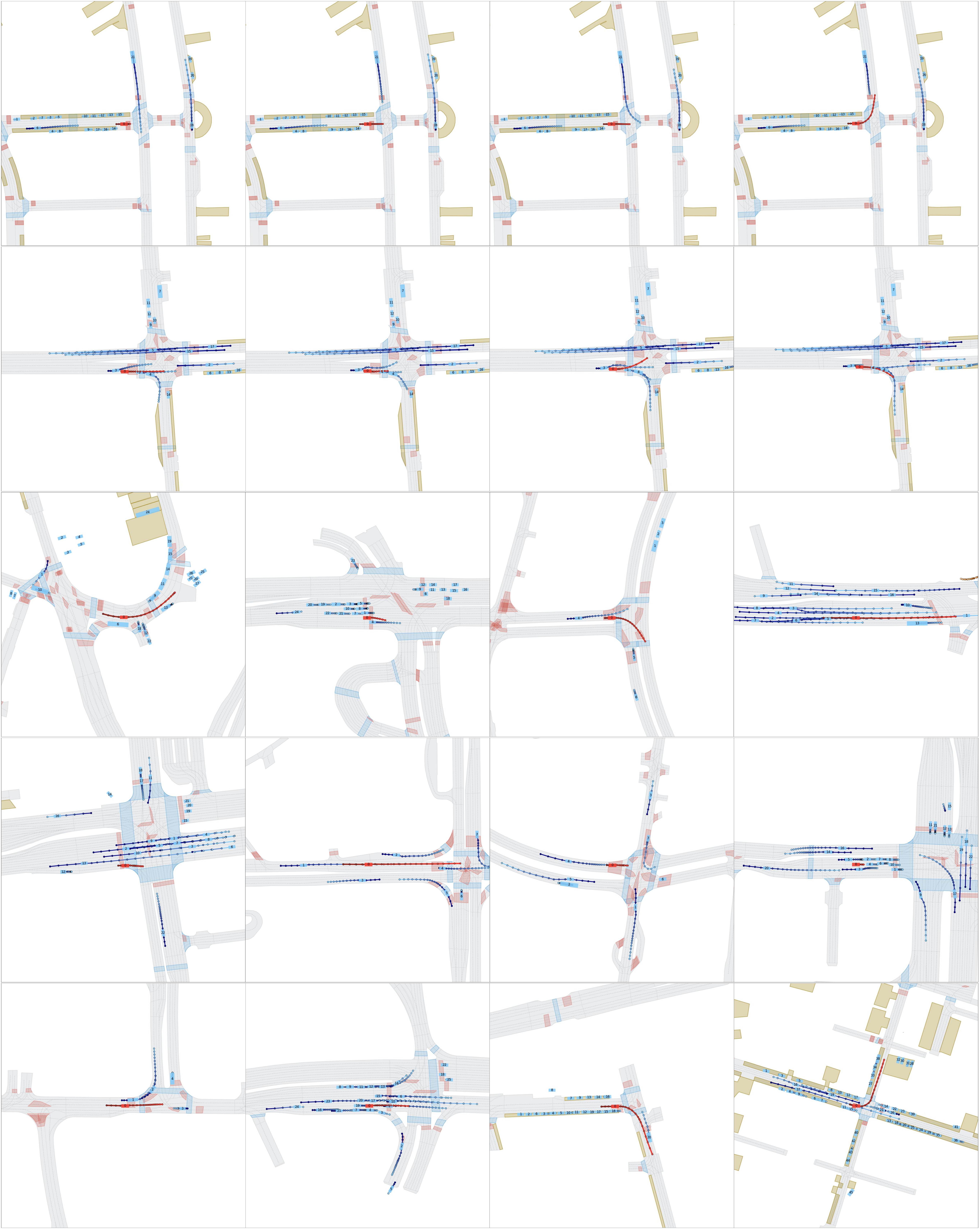}
    \caption{\textbf{Free Exploration on nuPlan.}
\textbf{Top (two rows):} Multiple inference runs on the same initial scene produce diverse scene variations.
\textbf{Bottom (three rows):} Additional examples across varied nuPlan scenarios, illustrating the breadth of freely generated scene variations.}
    \label{fig:Sup_norm}
\end{figure*}

\begin{figure*}[p]
    \centering
    \includegraphics[width=\linewidth]{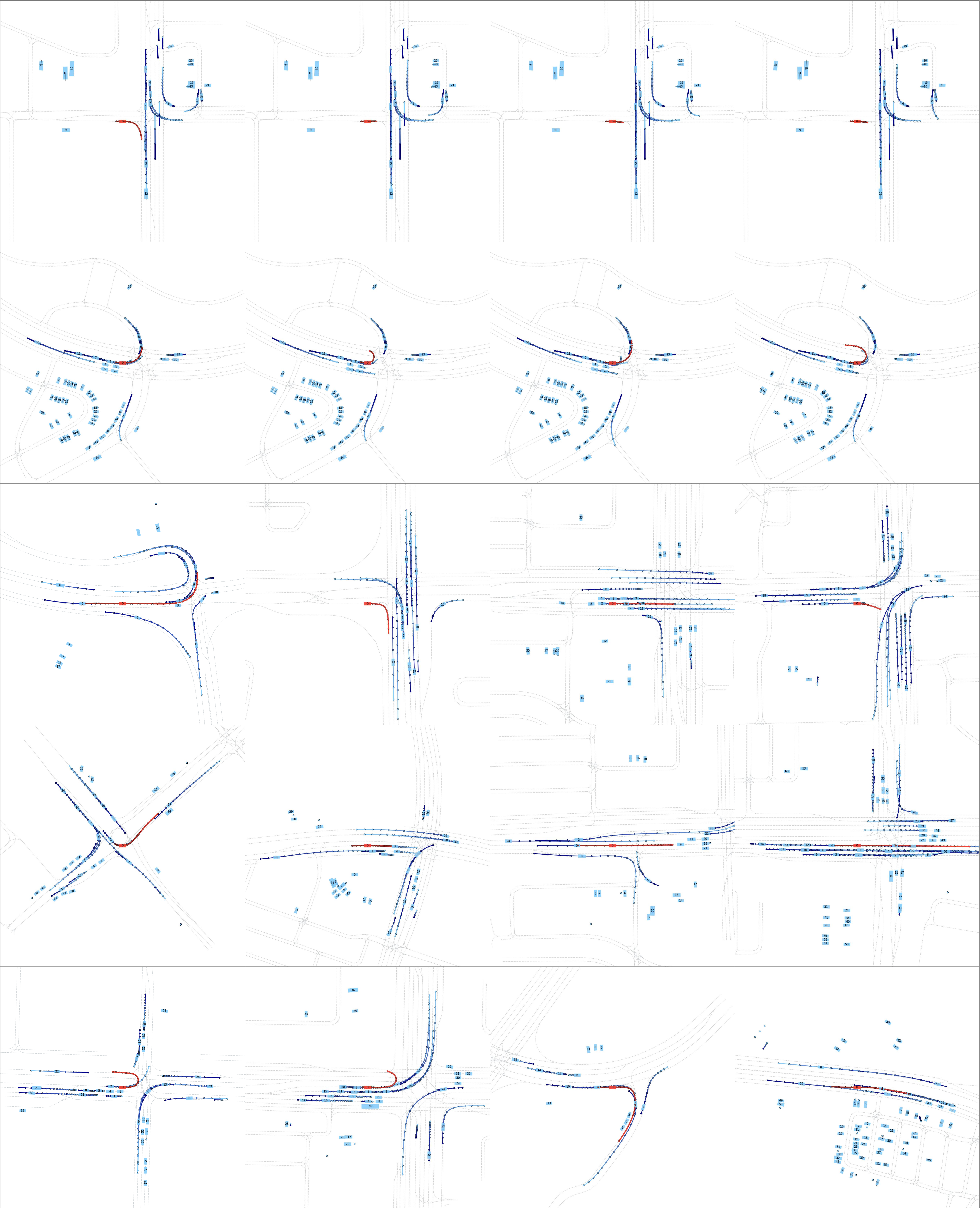}
    \caption{\textbf{Zero-Shot Free Exploration on Waymo.}
\textbf{Top (two rows):} Multiple inference runs on the same initial scene produce diverse scene variations.
\textbf{Bottom (three rows):} Additional examples across varied Waymo scenarios, illustrating the broad generalization ability and diversity of freely generated scene variations.}
    \label{fig:Sup_waymo}
\end{figure*}

\begin{figure*}[p]
    \centering
    \includegraphics[width=\linewidth]{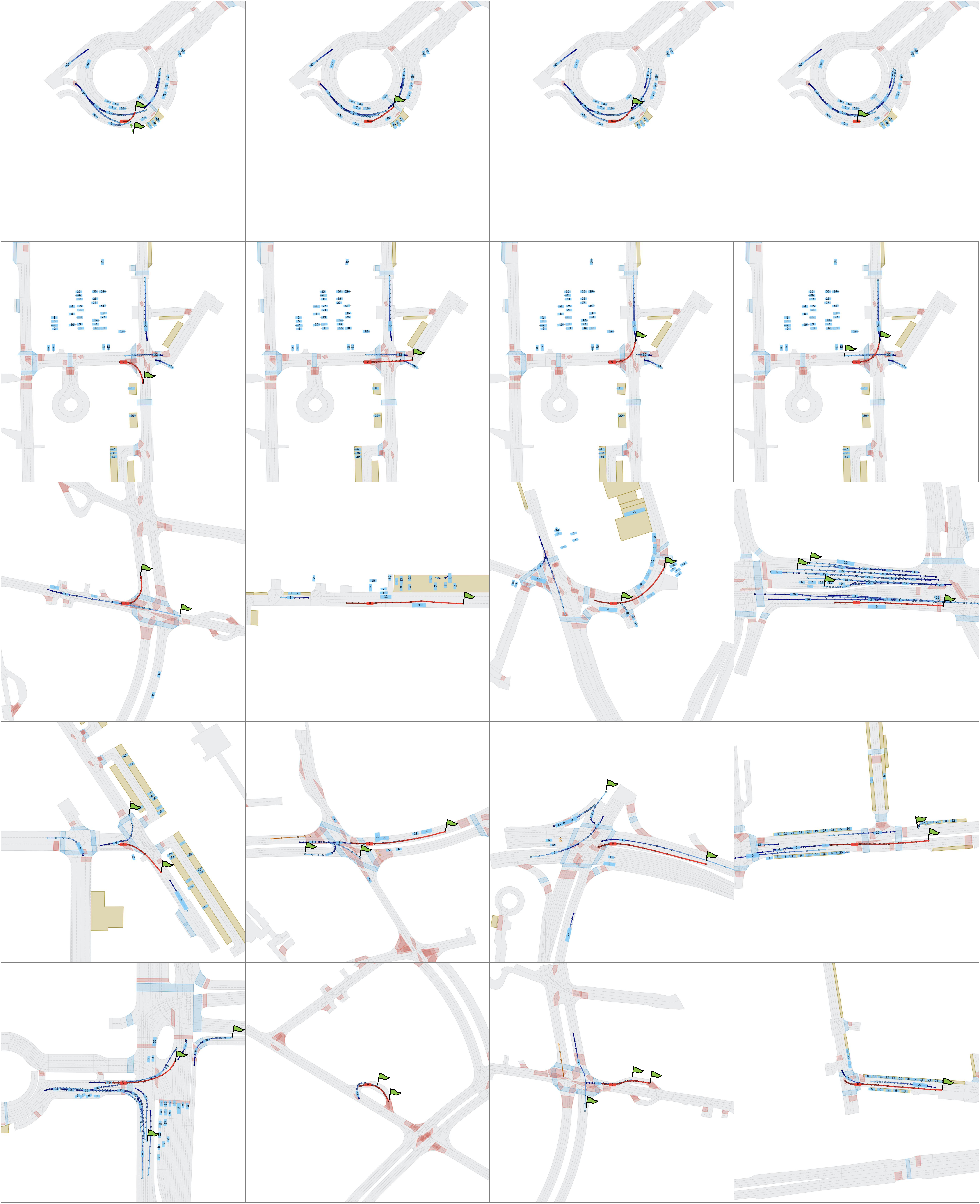}
    \caption{\textbf{Goal-Conditioned Scene Generation.}
\textbf{Top (two rows):} Starting from the same initial scene, manually specifying different goal points for selected agents yields diverse yet consistent goal-directed outcomes. The green flags indicate the user-defined goal locations.
\textbf{Bottom (three rows):} Additional examples across varied nuPlan scenarios, showcasing controllable and coherent behavior generation under different goal specifications.}
    \label{fig:gc}
\end{figure*}

\begin{figure*}[p]
    \centering
    \includegraphics[width=\linewidth]{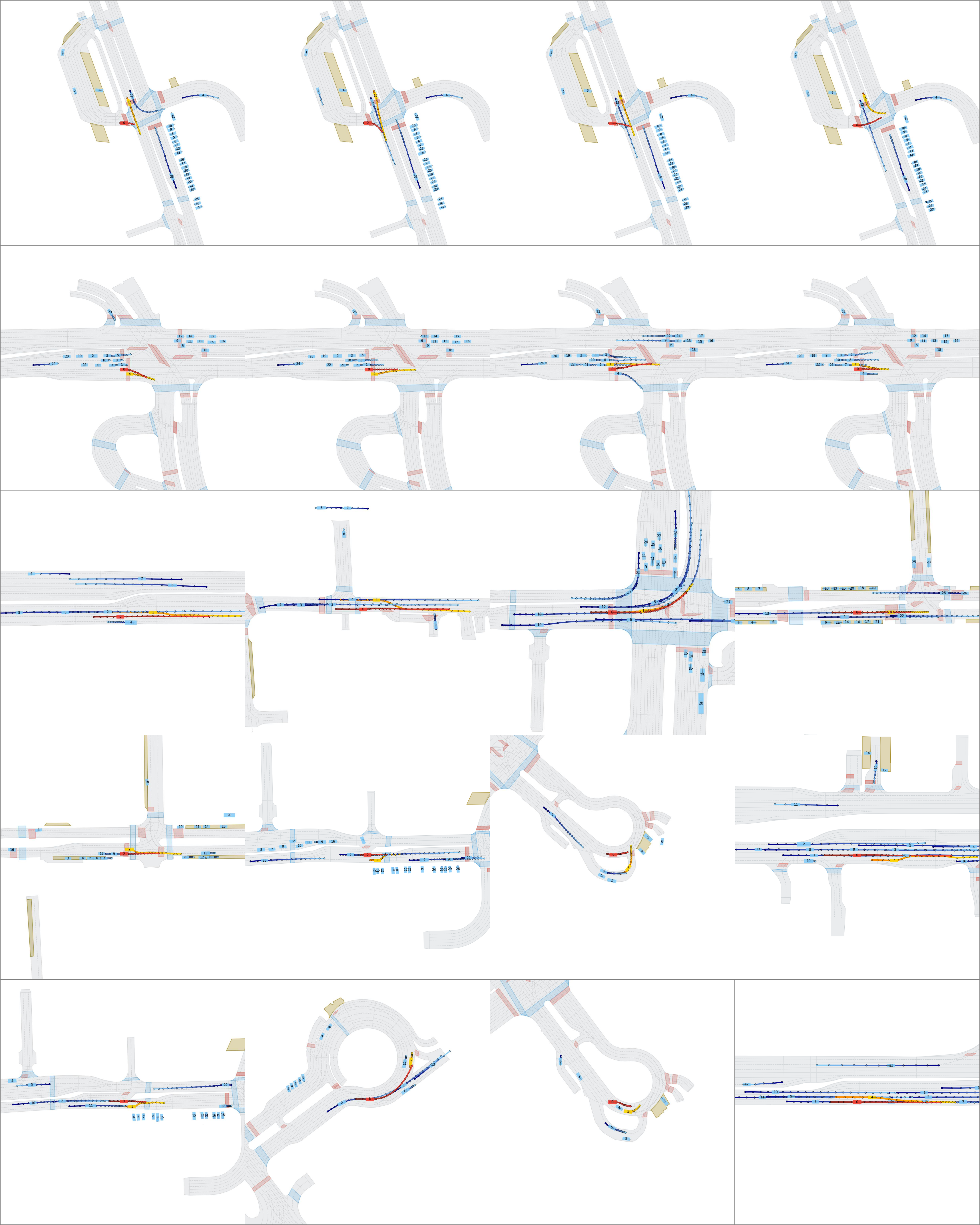}
    \caption{\textbf{Adversarial Scene Generation.}
\textbf{Top (two rows):} Varying attacker identities and multiple inference runs yield distinct adversarial outcomes from the same initial history.
\textbf{Bottom (three rows):} Additional examples across diverse scenarios and attacker types, where Nexus-$\mathrm{\Omega}_\mathrm{Adv}$ adapts to each attacker’s relative position and automatically generates contextually plausible attack behaviors}
    \label{fig:Sup_adv}
\end{figure*}

\begin{figure*}[!ht]
    \centering
    \includegraphics[width=\linewidth]{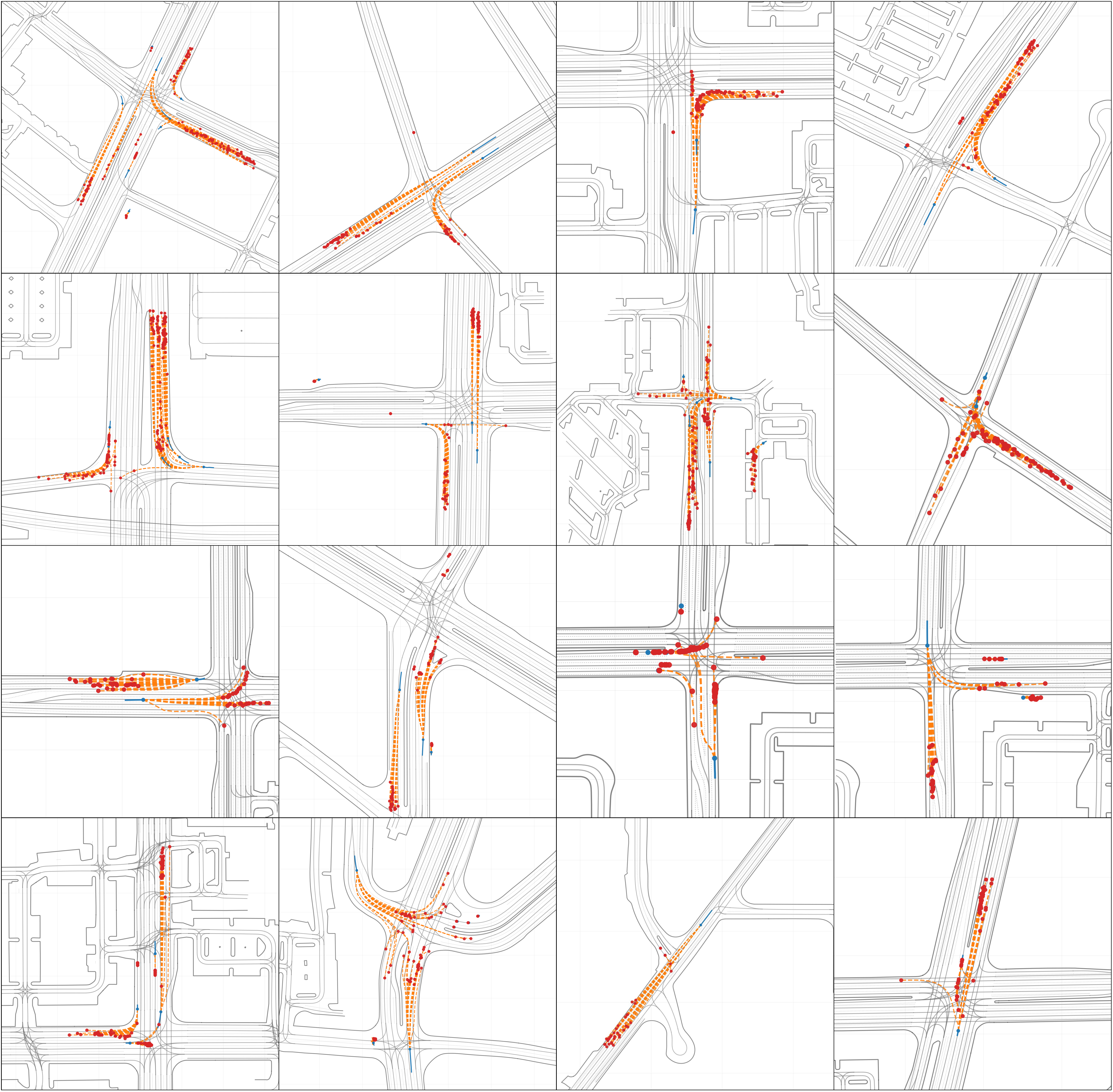}
    \caption{\textbf{Waymo Open Sim Agents Evaluation.} Thirty-two generated futures from the same historical initialization illustrate multimodal scene evolution while remaining physically plausible. Blue solid lines indicate historical trajectories with blue dots marking the current states, while orange dashed lines show inferred futures with red dots marking the final timestep.}
    \label{fig:Sup_wosac}
\end{figure*}

We provide extended qualitative visualizations to complement the
quantitative results. \cref{fig:Sup_norm}--\ref{fig:Sup_adv} and \cref{fig:Sup_wosac}
illustrate OMEGA's generation results across diverse road layouts, traffic densities, and task settings.

\paragraph{Free exploration on nuPlan}
\cref{fig:Sup_norm} shows additional qualitative results for
the free-exploration setting on nuPlan. For several scenes, we display
multiple stochastic samples generated from the same historical
initialization to illustrate the diversity of plausible futures
produced by OMEGA while preserving coherent multi-agent interactions.
We also include examples from a wide range of road geometries such as
straight roads, multi-lane segments, intersections, and merging areas,
each presented with one generated sample to highlight performance
across different map structures.

\paragraph{Free exploration on Waymo}
\cref{fig:Sup_waymo} provides complementary visualizations
for the zero-shot setting on Waymo. As with nuPlan, we show multiple
samples for a fixed scene together with a broader set of scenes that
highlight OMEGA's generalization to an unseen dataset with distinct
map geometries and traffic patterns.

\paragraph{Goal-conditioned generation}
\cref{fig:gc} presents extended results for
goal-conditioned controllability. For a fixed scenario, we display
multiple goal specifications for the ego and selected surrounding
vehicles, illustrating how OMEGA adapts trajectories to reach
distinct target states while maintaining interaction coherence.
Additional scenes across different road structures further show
consistent goal satisfaction and physically plausible multi-agent
coordination.

\begin{figure*}[!t]
    \centering
    \includegraphics[width=\linewidth]{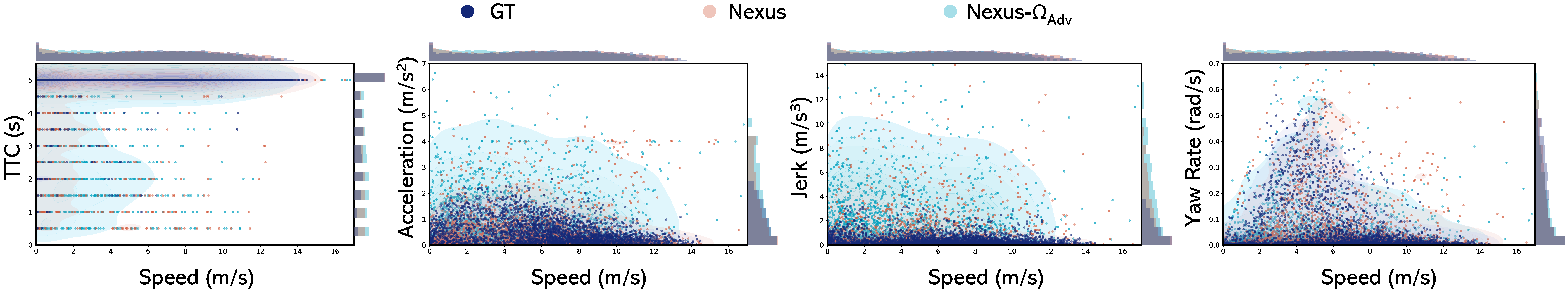}
    \caption{
        \textbf{Ego-kinematic distributions in adversarial scenarios.}
        Joint distributions of ego speed with TTC, acceleration, jerk,
        and yaw rate for ground truth (GT), Nexus, and
        Nexus-$\mathrm{\Omega}_{\mathrm{Adv}}$.  
        Marginal histograms use a logarithmic scale to emphasize rare
        events. Nexus-$\mathrm{\Omega}_{\mathrm{Adv}}$ produces more short-TTC
        frames, shifts ego speeds toward lower and medium ranges, and
        broadens acceleration and jerk distributions, indicating
        stronger adversarial effects on ego motion.
    }
    \label{fig:adv_dist}
\end{figure*}

\paragraph{Adversarial scenario generation}
\cref{fig:Sup_adv} presents qualitative results for adversarial
scenario generation. For the same historical initialization, selecting different vehicles as the designated attacker leads to distinct adversarial outcomes. These variations arise from
OMEGA$_\mathrm{Adv}$ adapting its refinement to the motion context of the chosen attacker and synthesizing attack behaviors that are consistent with its relative positioning and traffic
conditions. Additional examples cover a wide range of challenging
patterns such as emergency braking, merge-in maneuvers, cut-in events, and close-approach interactions, demonstrating that the generated scenes capture rich safety-critical behaviors while remaining aligned with realistic multi-agent dynamics. \cref{fig:adv_dist} shows the joint distributions of ego speed with TTC, acceleration, jerk, and yaw rate. Compared with Nexus and ground-truth logs, Nexus-$\mathrm{\Omega}_{\mathrm{Adv}}$ produces more short-TTC events at low-to-moderate speeds, shifts ego speeds toward lower values, and exhibits broader acceleration and jerk variations across speeds, indicating more diverse ego-motion patterns than natural data. Its comparable yaw-rate distribution suggests that the increased adversarial intensity is primarily longitudinal while preserving physically plausible dynamics.

\paragraph{Waymo Open Sim Agents evaluation}
\cref{fig:Sup_wosac} presents qualitative results under the Waymo Open Sim Agents evaluation. For each scene, we visualize 32 generated futures from the same historical initialization, illustrating diverse and multimodal scene evolutions while remaining physically plausible.

\subsection{Failure cases.}

\begin{figure*}[!ht]
    \centering
    \includegraphics[width=\linewidth]{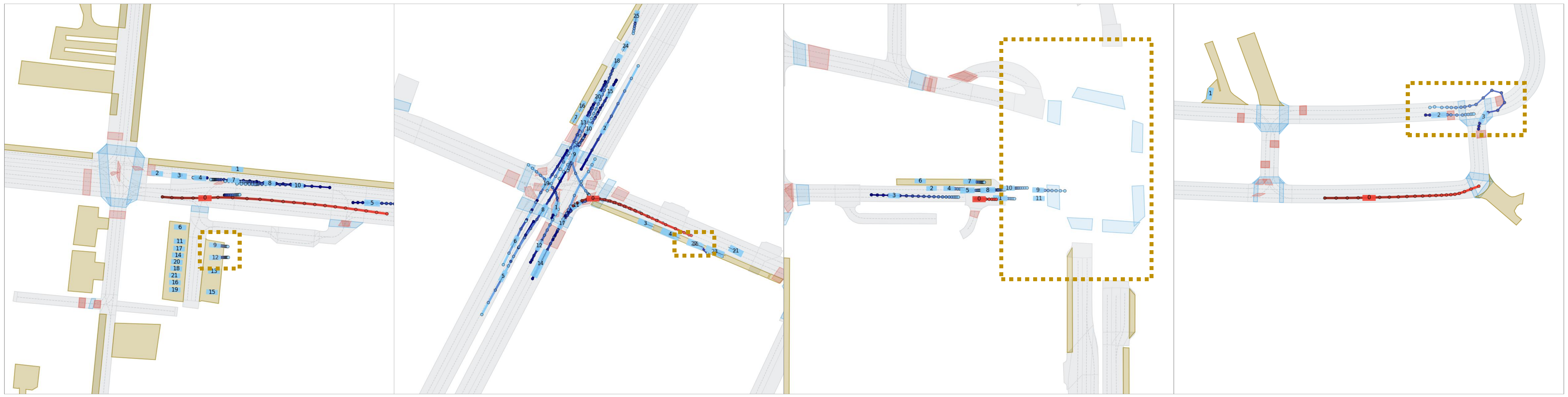}
    \caption{\textbf{Failure cases of guided scene generation.}}
    \label{fig:Sup_fail}
\end{figure*}

\cref{fig:Sup_fail} illustrates several situations in which the guided
generative process produces suboptimal behaviors. In the first example,
although the two-phase schedule improves long-horizon stability, tthe Rolling-Zero refinement remains
autoregressive over physical time. As a result, small per-step
errors can accumulate, producing minor autoregressive drift in
stationary agents. This issue can be mitigated by freezing
truly static background vehicles during sampling or by lightly
finetuning the backbone model under the Rolling-Zero schedule.

The second case stems from imperfections in the history conditioning of
real-world logs, where two agents may already overlap at the initial
state due to perception noise. Since the generative process inherits
this initialization, such collisions cannot always be avoided. This can
be addressed by filtering out scenarios in which the history contains
overlapping agents before generation.

A third situation arises when map information is incomplete or missing.
In these cases, map-aware components of the guidance become inactive and
the refinement relies solely on motion cues, which may lead to
off-road predictions. Ensuring consistent map availability or
incorporating fallback geometric priors can help alleviate this
behavior.

The fourth example illustrates that guided sampling operates within
the representational capacity of the underlying diffusion prior.
When the model proposes highly unrealistic trajectories during
denoising, the guidance step may not fully restore physical
consistency. Improving the backbone model therefore remains an
effective direction for future enhancement.


\end{document}